\documentclass[journal]{IEEEtran}
\usepackage{cite}
\usepackage{amsmath,amssymb,amsfonts}
\usepackage{graphicx}
\usepackage{textcomp}
\usepackage{xcolor}
\usepackage{algpseudocode}
\usepackage{algorithm}
\usepackage[hidelinks]{hyperref}
\def\BibTeX{{\rm B\kern-.05em{\sc i\kern-.025em b}\kern-.08em
    T\kern-.1667em\lower.7ex\hbox{E}\kern-.125emX}}

\newcommand{\bmat}[1]{\begin{bmatrix}#1\end{bmatrix}}

\begin{document}
\title{Stability and Transparency in Mixed Reality Bilateral Human Teleoperation}

\author{\IEEEauthorblockN{David Black,}
\and
\IEEEauthorblockN{Septimiu Salcudean}
}

\maketitle

\begin{abstract}
Recent work introduced the concept of human teleoperation (HT), where the remote robot typically considered in conventional bilateral teleoperation is replaced by a novice person wearing a mixed reality head mounted display and tracking the motion of a virtual tool controlled by an expert. HT has advantages in cost, complexity, and patient acceptance for telemedicine in low-resource communities or remote locations. However, the stability, transparency, and performance of bilateral HT are unexplored. In this paper, we therefore develop a mathematical model and simulation of the HT system using test data. We then analyze various control architectures with this model and implement them with the HT system to find the achievable performance, investigate stability, and determine the most promising teleoperation scheme in the presence of time delays. We show that instability in HT, while not destructive or dangerous, makes the system impossible to use. However, stable and transparent teleoperation are possible with small time delays ($<200$ ms) through 3-channel teleoperation, or with large time delays through model-mediated teleoperation with local pose and force feedback for the novice.
\end{abstract}


\section{Introduction}
Many remote and underresourced communities experience severe challenges in accessing qualified medical care. For example, ultrasound imaging is important, widely used, and much lower cost than other modalities such as CT or MR. However, capturing and interpreting ultrasound images requires a high degree of expertise that is not commonly present in many small communities. As a result, a sonographer or radiologist must be transported to the town on a regular basis, or patients must be sent to a major medical center. Either case leads to long wait times and difficulty handling urgent cases. In communities across Canada, patients are flown hundreds of kilometers for standard ultrasound exams. This takes up to three days and exerts a high social and financial cost on the community. 

Therefore, tele-ultrasound is an important and growing field. However, current commercially available technologies are often impractical. Video teleguidance is simple, low-cost, and accessible to anyone but is highly inefficient and imprecise if the person being guided does not already have ultrasound experience \cite{black2023hci}. On the other hand, robotic teleultrasound gives the physician complete and precise control but is expensive and complex to set up and maintain. We thus recently introduced a novel teleguidance method called human teleoperation to address the shortcomings of both existing approaches \cite{black2023hci,black2024cag}. This method is also applicable to many other remote guidance applications.

In human teleoperation, a local novice, the ``follower", performs an ultrasound exam on a patient while being guided by a remote operator, the sonographer or radiologist. The follower wears a mixed reality (MR) head-mounted display (HMD) which projects a virtual ultrasound probe into their field of view. The pose of this virtual probe is controlled in real time by the operator, who manipulates a haptic device with an ultrasound probe-shaped end effector. As the operator moves the virtual probe, the follower aligns their real probe to the virtual one and follows it as it moves, thus achieving the operator's desired motion. The ultrasound image and video from the follower's HMD are streamed to the operator so they can carry out the procedure. The two sides are also in verbal communication. The tracking of the MR virtual probe is very intuitive, and no prior ultrasound experience is required for the follower. Indeed, the ability of the follower to track the MR input was previously characterized \cite{black2023ijcars,black2024tmrb}, showing tracking lags of 200-300 ms and steady state error of less than 3 mm. Such tight coupling of the operator and follower enable this teleguidance system to be designed and analyzed like a robotic teleoperation system, leading to the name human teleoperation. 

While the position and orientation of the ultrasound probe are key during procedures, the applied force is also an important factor. Applying the correct level of force avoids excessive deformation of structures like blood vessels, displaces gas which otherwise obstructs the image, and allows imaging below the ribs, for example. Therefore, the ultrasound probe in human teleoperation is instrumented with force sensing, as previously described \cite{black2023whc, black2024ft}. Additionally, the probe pose is tracked visually by the HMD \cite{black2024pose} to transform the force to known coordinates. Furthermore, the haptic device of the operator has three actuated joints, allowing it to apply forces to the operator's hand. In this way, it is possible to render a realistic haptic sense of the patient to the operator.

Haptic feedback is important in teleoperation. In robot-assisted minimally invasive surgery \cite{saracino2019} and surgical training \cite{van2009}, haptic feedback has been shown to increase task performance and decrease tissue damage \cite{alleblas2017,tholey2005}. Similarly, in teleoperation of mobile robots \cite{diolaiti2002}, unmanned aerial vehicles \cite{lam2009}, and robots for micro-manipulation and assembly \cite{bolopion2013}, haptic feedback has shown utility. In human teleoperation for teleultrasound, there are three primary reasons to use haptic feedback. First, sonographers are accustomed to resting their hand on the patient and exerting relatively large forces up to approximately 20~N \cite{smith2003}. Without this support, the experience feels unrealistic and unintuitive. More importantly, it is very difficult to hold the probe at exactly the height of the patient surface or to make precise motions if the operator's hand is floating in space instead of resting on the surface. Finally, the operator uses their sense of touch to perform the exam, for example by palpating tissue or by feeling the ribs so as to image between them. Therefore, haptics is essential in human teleoperation.

From the perspective of teleoperation literature, haptic feedback increases the transparency of the teleoperation. Transparency describes how well the follower and operator sides match each other in terms of force and position (velocity) \cite{lawrence1993}. In a perfectly transparent system, the velocity and force of the operator and follower (commonly referred to in prior literature as master and slave, respectively) would be exactly equal at all times. Thus, the follower would perfectly track the operator's motion and the operator would feel exactly as if they were touching the remote environment directly. This is a primary goal in teleoperation that has been addressed using many approaches including direct force feedback \cite{burdea1991}, position feedback \cite{kim2005}, combined position and force feedback \cite{zhu1999, hashtrudi2000}, impedance feedback \cite{salcudean1999}, rate control \cite{mobasser2008}, adaptive control \cite{lee1998}, local force feedback at the follower \cite{hashtrudi1999},  and more.

The other essential objective in teleoperation is stability. Especially in medical applications in which the follower robot interacts with patients, instability in the system can constitute a health hazard. Unfortunately, even a small amount of time delay can destabilize nominally stable bilateral teleoperation systems \cite{sheridan1993}. Since any remote teleoperation system inevitably includes communication at a distance, leading to delays, this is an important problem that has been studied extensively. Due to the complex, nonlinear nature of teleoperators, system passivity is commonly used to show stability \cite{nuno2011}. For example, it has been shown that transmitting wave variables rather than the values themselves guarantees the passivity of the system under arbitrary time delays \cite{niemeyer1991, aziminejad2008}. However, this can degrade tracking accuracy and transparency. Thus, others have used predictive methods such as Smith predictors \cite{smith2003,smith2005,smith2006} and model predictive control \cite{sirouspour2006} to eliminate the time delay by predicting the feedback before it arrives. Similarly, model-mediated control allows instantaneous feedback by keeping a local model of the remote environment on the operator side \cite{xu2016}. As these methods are affected by modeling accuracy, however, others have studied transmitting adaptive combinations of velocity and force \cite{zhu2000}, and robust control approaches \cite{haddadi2012robust}. A particularly successful method called time-domain passivity was proposed by Hannaford and Ryu \cite{hannaford2002, ryu2005}, in which an energy-like quantity of the system is observed and dissipative elements are added to eliminate only the net energy output, thus maintaining passivity with less degradation of performance.

Despite the extensive investigation of robotic teleoperation, no research exists that examines the stability and transparency of human teleoperation. In human teleoperation, the slave robot is replaced by a human, leading to fundamentally different system behavior. In particular, the human follower may be considered passive at all times \cite{hogan1987}, but their reaction time and imperfect tracking lead to potentially larger delays and inaccurate motions and forces. The follower can intrinsically respond to changing environment conditions and does not require trajectory planning around joint limits and singularities, nor initial calibration, safe deceleration, or compliant controllers. However, they must be guided effectively according to the capabilities of their vision and perception, without providing too much input so as to cause cognitive overload or rapid fatigue. Thus, human teleoperation is influenced by other factors than robotic teleoperation, but stability and transparency remain paramount. To achieve efficient, performant teleoperation, it is thus important to study the stability and transparency of the human teleoperation system. 

To this end, this paper presents an initial application of the concepts of stability and transparency from telerobotics to bilateral human teleoperation. In particular, the following contributions are included:
\begin{itemize}
    \item A physical model of the human-in-the-loop system is derived and used to create a hybrid matrix formulation of the teleoperation with time delays in Section \ref{sec:2b}.
    \item This model is used with several candidate controllers to investigate their respective transparency and expected performance for human teleoperation in Section \ref{sec:2c}.
    \item In Section \ref{ssec:stab}, the stability of bilateral human teleoperation is investigated.
    \item In Section \ref{ssec:rob}, the robustness of the derived models to parametric and dynamic uncertainty is explored using the structured singular value. 
    \item Finally, limited tests with the real system and volunteer followers were performed to validate the modeling. The setup is described in Sections \ref{sec:2f} and \ref{sec:2g}, and the results are presented in Section \ref{sec:res}.
\end{itemize}
While this paper does not constitute a comprehensive analysis or test of all possible human teleoperation approaches, it lays the foundations both for future practical tests with human volunteers and for further algorithmic development, for example in predictive or robust control schemes.

\section{Methods}
\subsection{Human Teleoperation}\label{sec:2a}
\begin{figure*}[t]
    \centering
    \includegraphics[width=0.8\linewidth]{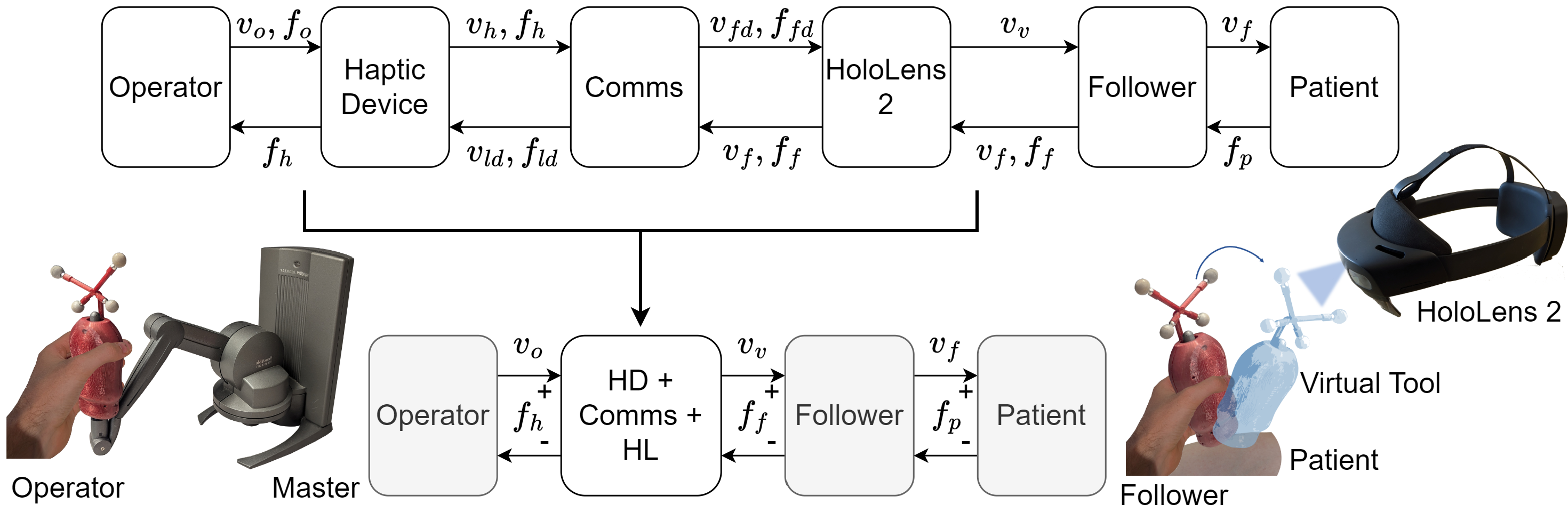}
    \caption{Conceptual overview of the teleoperation system. The gray boxes of operator, follower, and patient have to be modeled while the haptic device, communications, and HoloLens 2 represent the part of the system we can directly control.}
    \label{fig:overallDiagram}
\end{figure*}
Human teleoperation enables precise, tightly-coupled remote guidance of a person's hand motion by an operator. A block diagram is shown in Fig. \ref{fig:overallDiagram}. The system consists of three primary components: the operator side, follower side, and communication. On the follower side, the follower wears an MR headset, the HoloLens 2 (Microsoft, Redmond, WA), and holds a manual tool that is instrumented with force sensing. Most work so far has focused on remote ultrasound, so the tool is an ultrasound probe with a force sensing shell \cite{black2023whc,black2024ft}. The probe also has infrared markers that are tracked by the HoloLens, giving access to the probe's pose in real time \cite{black2024pose}. The operator side consists of a monitor on which the ultrasound image and video stream of the follower are displayed, as well as a haptic device (TouchX, 3DSystems, Rock Hill, SC) that the operator uses to input their desired motion. This motion is executed contemporaneously by a virtual ultrasound probe displayed in the HoloLens 2. The follower aligns their ultrasound probe with the virtual one to match the desired pose of the operator. The measured force and pose are fed back to the operator, where they can be used to generate forces in the haptic device.

The communication is performed over the Internet through WebRTC, a fast peer-to-peer framework, as described previously \cite{black2024tmrb}. Prior work has also examined the ability of people to track mixed reality inputs, analyzing the tracking error, steady-state error, and lag of their motions and forces compared to the operator's input \cite{black2023ijcars}. Step and frequency response tests were also carried out \cite{black2024tmrb}. The results of these tests are used to inform the mathematical modeling in the following section.

\subsection{Modeling}\label{sec:2b}
\begin{figure}[h]
    \centering
    \includegraphics[width=\linewidth]{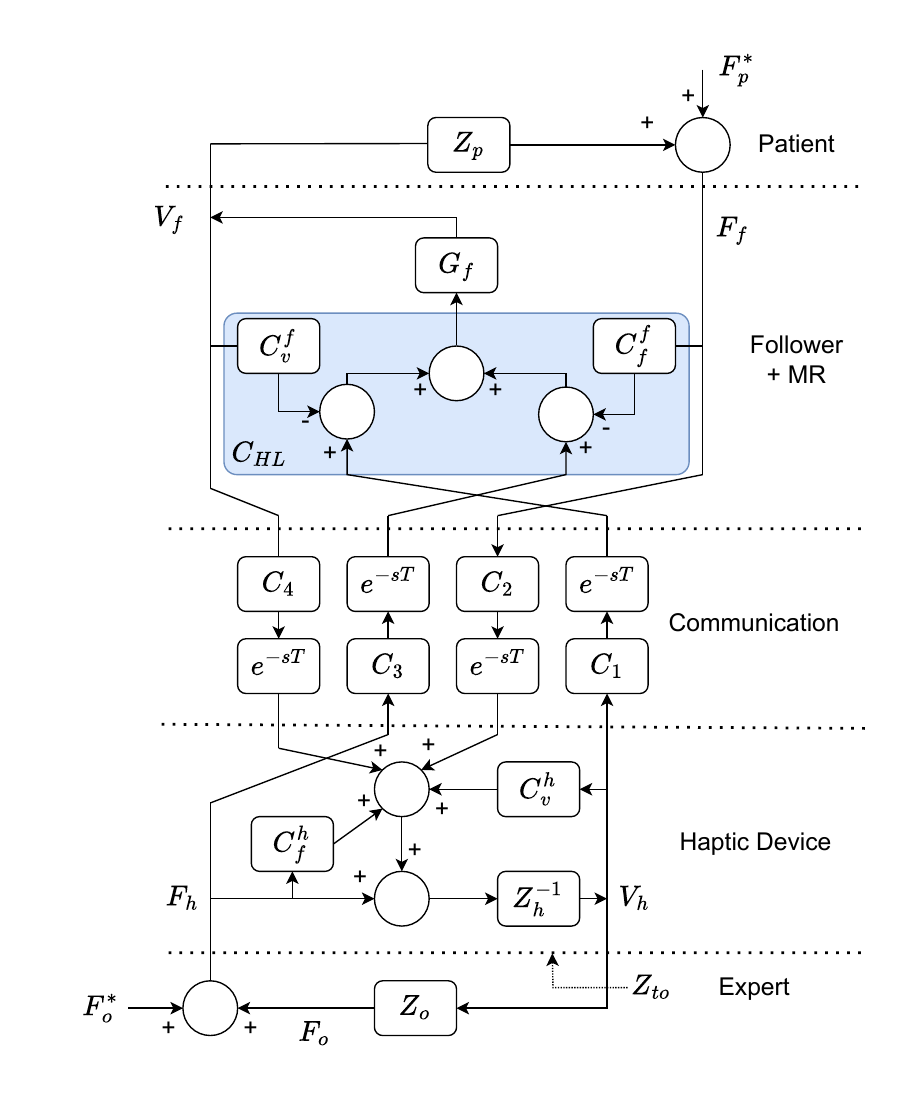}
    \caption{A general model of human teleoperation. The operator (subscript $o$) interacts with a haptic device (subscript $h$) while the follower ($f$) interacts with the patient ($p$). The communication channel induces time delays of $T$ on the force and velocity which are transmitted bilaterally. The MR headset creates a visual control output potentially using all four channels of force and velocity, denoted $C_{HL}$. The follower responds to the MR input according to the follower transfer function, $G_f$. The controllers $C^i_j$ are on the follower or haptic device respectively with $i=f$ or $h$, and involve force or velocity respectively with $j=f$ or $v$.}
    \label{fig:teleop}
\end{figure}

A general model of human teleoperation is shown in Fig.~\ref{fig:teleop}. We consider a one-dimensional, linear time-invariant (LTI) system which can be extended to three dimensions accordingly. Other than the follower side, the structure is similar to previous work in teleoperation. Indeed, human teloperation differs from robotic teleoperation primarily in the response of the follower compared to the slave robot. While both follow the commanded motion and/or force and consequently interact with the patient, at a low level the robot is controlled through joint torque commands and must explicitly account for its own dynamics. Conversely, the follower is controlled through position commands by following the virtual tool and implicitly accounts for his/her own arm dynamics. Furthermore, the robot, depending on the controller, can become unstable. On the other hand, the human follower is always stable and can be treated as a passive system \cite{hogan1987}. Thus, the stability analysis is simplified. However, the follower's response is slower than most robots' and less accurate, with stochastically varying performance base on focus and other factors, so achieving transparency is more difficult.

Beyond the virtual tool pose, it is possible to give additional input to the follower such as changing the tool color or displaying an error bar, arrow, or visual renderings in the MR headset. These options could be added in the blue $C_{HL}$ box in Fig. \ref{fig:teleop}. How to model the human response to color or error bars mathematically is, however, an open question. Furthermore, we found previously \cite{black2023ijcars} that further rendering is distracting and leads to cognitive overload for the follower. Hence, this paper examines only the virtual tool pose and other renderings are left for future work.

Given these assumptions about the human teleoperation system, we develop models of the follower and operator in contact with the patient and haptic device respectively, and study how these can be interconnected in a bilateral teleoperation system.
\begin{figure}[h]
    \centering
    \includegraphics[width=0.6\linewidth]{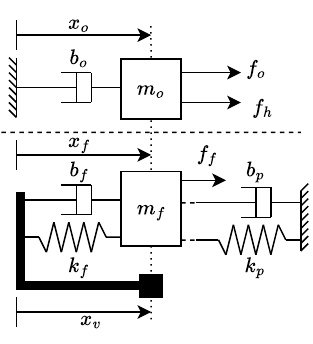}
    \caption{Models of the operator and haptic device (top), and the follower potentially in contact with the patient (bottom). The damping of the operator arm and haptic device together is represented by $b_o$. The black fixture attached to the follower mass is rigid and massless. The square end at position $x_v$ represents the virtual tool. The follower force $f_f=-k_px_f-b_p\dot{x}_f$ cancels the patient dynamics at all times. Thus, when the follower motion has converged (i.e. $b_f$ and $k_f$ are at equilibrium), the follower position $x_f$ should equal $x_v$.}
    \label{fig:models}
\end{figure}
\subsubsection{Follower}
The follower model was derived from measured response data of 11 volunteer followers from previous studies \cite{black2023ijcars,black2024tmrb}. It was found that a second-order system with two poles and one zero matched the measured behavior well. This is equivalent to attaching the follower's hand and tool, which have a certain mass, to the virtual tool by a spring and damper, as shown in Fig. \ref{fig:models}. The follower either moves freely in space or is in contact with the patient, represented by an impedance $Z_p=b_ps+k_{p}$, which is taken to be constant.

We assume the follower applies whatever force is necessary to match the virtual tool, whether or not they are in contact with the patient, so there is no steady-state error. In other words, $f_f=-k_px_f-b_p\dot{x}_f$ at all times, so it perfectly cancels the patient contact dynamics. In this way, intermittent contact with the patient has no effect on the follower's stability or tracking, which is as expected from experience; the human hand does not become unstable, even in high-frequency contact with a stiff surface. The follower can thus be represented by the passive (strictly passive if $b_f>0$) transfer function
\begin{align}\label{eqn:foltf}
    x_f&=\left(\dfrac{b_fs+k_f}{m_fs^2+b_fs+k_f}\right)x_v = G_fx_v\\
    f_f&=-(b_ps+k_p)x_f
\end{align}
where $k_{p}$, $b_p$ are zero when not in contact with the patient. For this to be LTI, we assume the patient is unmoving and their impedance is constant. This is approximately true when scanning only one region, for example the abdomen. An advantage of this model is that we do not have to consider switching contact, which is handled implicitly by the follower.

\begin{figure*}[t]
    \centering
    \includegraphics[width=\linewidth]{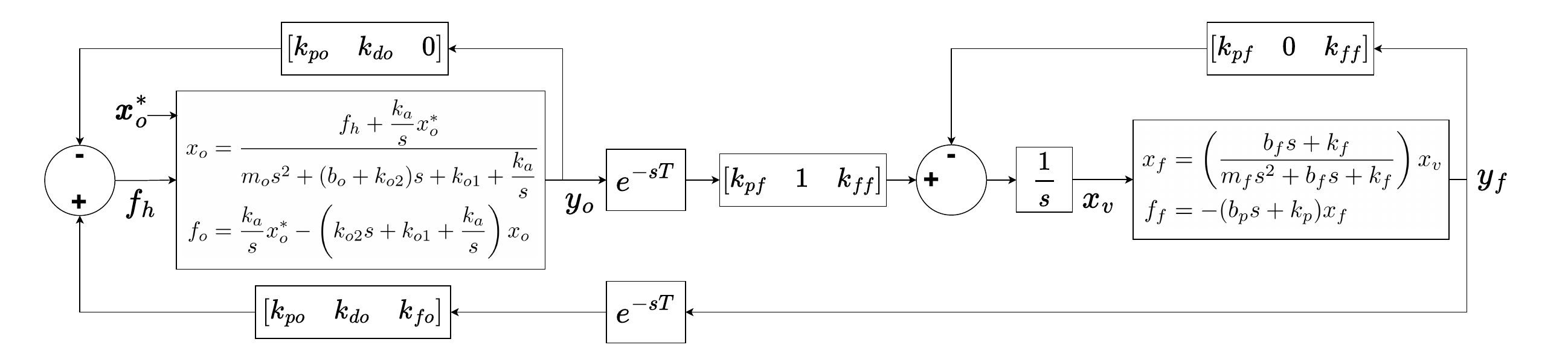}
    \caption{Interconnection of the derived models in the closed-loop teleoperation system. Different behavior is achieved by changing the feedback and feed-forward gains of the positions, velocities, and forces. The operator-side model is given in Equation \ref{eqn:exptf} while the follower side is in Equation \ref{eqn:foltf}.}
    \label{fig:tfTeleop}
\end{figure*}

\subsubsection{Operator and Haptic Device}
The operator hand holding the haptic device is modeled as a mass and damper system which is subject to the applied force of the operator and the haptic device, as shown in Fig. \ref{fig:models}.
\begin{equation}
    m_o \ddot{x}_o + b_o\dot{x}_o = f_h+f_o
\end{equation}
The operator changes their applied force depending on the haptic device force, to track a desired trajectory. In particular, to choose a representative response, we use a proportional-integral-derivative (PID)-inspired controller, choosing
\begin{equation}
    f_o=-k_{o1}x_o-k_{o2}\dot{x}_o+k_a\int(x_o^*-x_o)dt
\end{equation}
where $x_o^*$ is the operator's desired motion. These gains can be tuned to obtain a response that resembles the recorded operator data. We ignore the operator's desired velocity for simplicity as it introduces a zero eigenvalue and the operator is primarily interested in position. We thus find the following representation of the operator
\begin{align}\label{eqn:exptf}
    x_o&=\dfrac{f_h+\dfrac{k_a}{s}x_o^*}{m_os^2+(b_o+k_{o2})s+k_{o1}+\dfrac{k_a}{s}}\\
    f_o &= \dfrac{k_a}{s}x_o^*-\left(k_{o2}s+k_{o1}+\dfrac{k_a}{s}\right)x_o
\end{align}
The state space formulations of the operator and follower are derived in the Appendix, where we show that the expert controller can be framed as servo control and tuned using pole placement.

\subsubsection{Teleoperation System}
The full teleoperation system is shown in Fig. \ref{fig:tfTeleop}, given the models derived above. Let $y_j=\bmat{x_j&\dot{x}_j&f_j}^\top~j=\{o,f\}$ be the output of the operator and follower, respectively. We assume the follower sees only the virtual tool pose and the operator receives feedback only by forces applied to the haptic device. This does not take into account the operator's visual feedback from video or other sensor streams such as ultrasound images, nor the verbal communication between operator and follower. In this case, we control the virtual tool pose, $x_v$, for the follower, and the haptic device force, $f_h$ for the operator. Each can be a function of the position, velocity, and force of the operator and follower as shown in the diagram. Some gains are never used, so they are set to zero. For example, the haptic device is not equipped with a force sensor, so its actual force cannot be fed back. Thus, the general inputs to the operator and follower are
\begin{align}
    f_h &= k_{fo}f_f + k_{po}(x_f-x_o) + k_{do}(\dot{x}_f-\dot{x}_o)\label{eqn:lpfLaw1}\\
    \dot{x}_v &= \dot{x}_o + k_{pf}(x_o-x_f) + k_{ff}(f_h-f_f)\label{eqn:lpfLaw}
\end{align}
Let $K_{oo}=\bmat{k_{po}&k_{do}&0}$, $K_{of}=\bmat{k_{pf}&1&k_{ff}}$, $K_{ff}=\bmat{k_{pf}&0&k_{ff}}$, and $K_{fo}=\bmat{k_{po}&k_{do}&k_{fo}}$. Then in terms of the individual model outputs, these expressions become
\begin{align}\label{eqn:fbe}
    f_h &= K_{fo}y_f - K_{oo}y_o\\\label{eqn:fbf}
    \dot{x}_v &= K_{ff}y_f - K_{of}y_o
\end{align}
For non-zero time delay, the equations are:
\begin{align}\label{eqn:fht}
    f_h &= e^{-sT}K_{fo}y_f - K_{oo}y_o\\\label{eqn:xvt}
    \dot{x}_v &= K_{ff}y_f - e^{-sT}K_{of}y_o
\end{align}
The full state space notation of the closed-loop teleoperation system is shown in the Appendix.

\subsubsection{Simulation}
A Simulink model was developed to simulate the system and test the controllers. This directly implemented the state-space models described in the previous subsections with some practical considerations. The haptic device force was set to saturate at 7 N, approximately the maximum force of the haptic device, and the environment was made approximately as stiff as a human abdomen during ultrasound \cite{guimaraes2020}. Next, the operator and follower parameters were independently estimated to obtain responses similar to previous measurements \cite{black2023ijcars,black2024tmrb}. Finally, the model and feedback gains were tuned with all components of the teleoperation connected together, to again match the expected performance. 

\begin{figure}[h]
    \centering
    \includegraphics[width=0.85\linewidth]{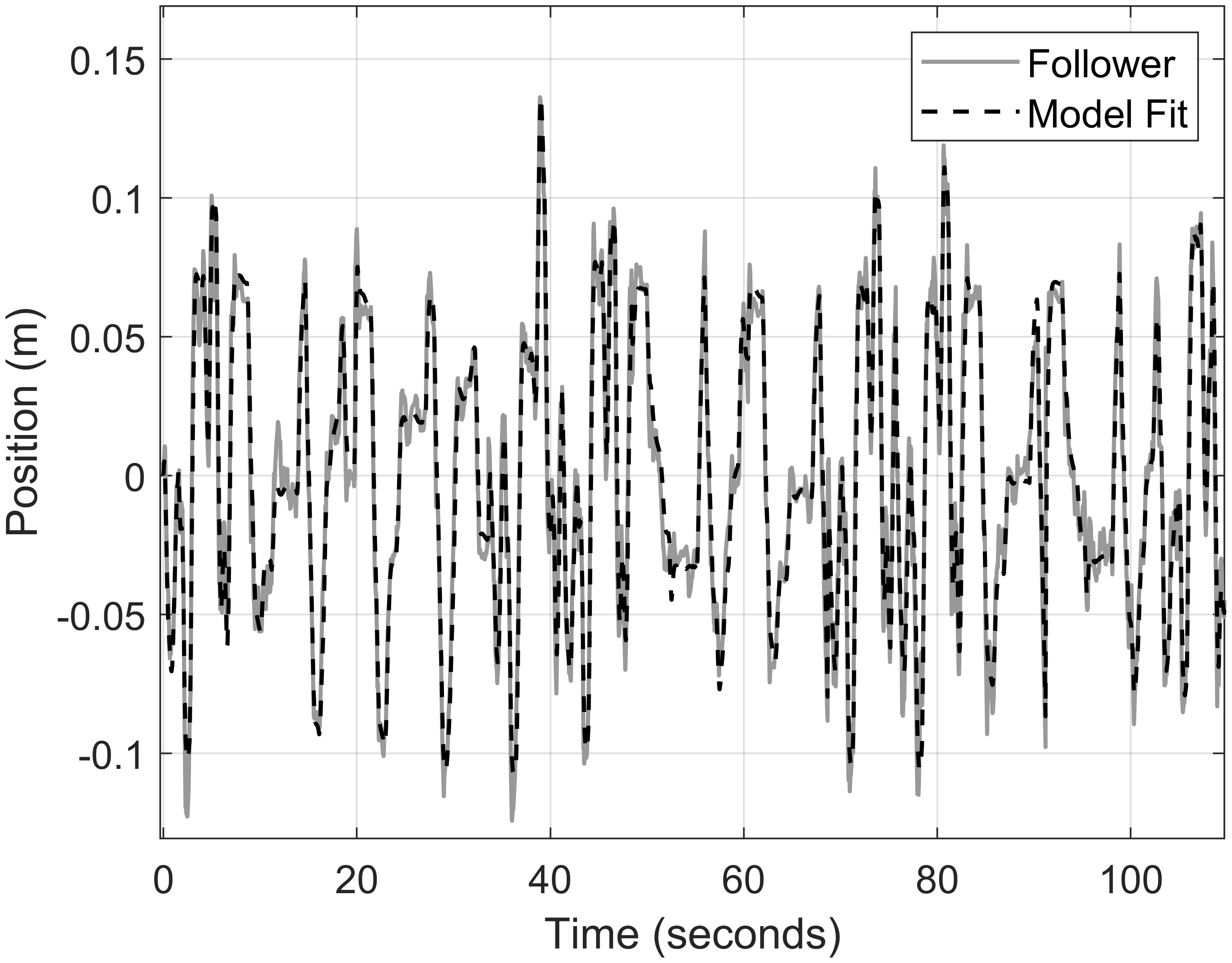}
    \caption{The derived model with parameters fitted to the measured data from previous tests \cite{black2023ijcars,black2024tmrb}. The MSE between the two is 0.67 mm.}
    \label{fig:modelFit}
\end{figure}

The chosen model parameter values were obtained using the MATLAB linear system identification toolbox and some manual tuning. Specifically, linear grey-box estimation was utilized with Adaptive subspace Gauss-Newton search to fit parameters to the derived ordinary differential equations, with non-negativity constraints and regularization on all parameters, and a stability constraint on the system. The final system fit the measured dataset with a mean-squared error (MSE) of 0.67 mm in the follower motion. The fit accuracy is relatively insensitive to changes in the parameters, which shows robustness in the model. The fit MSE has a partial derivative of 0.125 mm/kg for follower mass with all other parameters fixed at their chosen values. All other parameters have slopes of less than 0.0625 mm/unit in magnitude. The measured and fitted data are shown in Fig. \ref{fig:modelFit}.

In this way, approximate numerical values could be assigned to the various patient, follower, and operator parameters. These values and simulated models are useful for testing controllers before implementing them on physical hardware, and for numerically assessing the stability of a controller. An effective set of parameters is shown in Table \ref{tab:params}.

\begin{table}[h]
    \centering
    \caption{Approximate model parameter values determined through simulation and system identification based on previous data~\cite{black2023ijcars}. All units are N, m, s, kg.}
    \begin{tabular}{||c|c||c|c||c|c||}\hline
        \multicolumn{2}{||c||}{Patient} & \multicolumn{2}{c||}{Follower} & \multicolumn{2}{c||}{Operator} \\\hline
        $k_p$ & 10 & $k_f$ & 1 & $k_a$ & 100\\\hline
        $b_p$ & 1 & $b_f$ & 0.275 & $b_o$ & 0.1\\\hline
        $m_p$ & 0.02 & $m_f$ & 0.02 & $m_o$ & 0.1\\\hline
         &  &  &  & $k_{o1}$ & 0\\\hline
          &  &  &  & $k_{o2}$ & 80\\\hline
    \end{tabular}
    \label{tab:params}
\end{table}

\subsection{Transparency}\label{sec:2c}
Bilateral teleoperation systems are commonly expressed using a hybrid representation:
\begin{equation}
    \bmat{f_h\\-\dot{x}_f}  = \bmat{h_{11} & h_{12} \\ h_{21} & h_{22}}\bmat{\dot{x}_o\\f_f} = H(s)\bmat{\dot{x}_o\\f_f}
\end{equation}
For perfect transparency, $\dot{x}_o=\dot{x}_f$ and $f_h=f_f$, so the ideal hybrid matrix consists of elements $h_{11}=h_{22}=0$, $h_{12}=1$, and $h_{21}=-1$. Equations \ref{eqn:fht} and \ref{eqn:xvt} together with the operator and follower models can be expressed as a hybrid matrix with the following elements:
\begin{align}
    h_{11} &= \dfrac{e^{-s2T}(k_{po}+k_{do}s)\left(1-G_f^{-1}\right)}{G_f^{-1}s+k_{pf}-e^{-s2T}k_{ff}(k_{po}+k_{do}s)}\\
    h_{12} &= \dfrac{\left(G_f^{-1}s+k_{pf}\right)k_{fo}-k_{ff}(k_{po}+k_{do}s)}{G_f^{-1}s+k_{pf}-e^{-s2T}k_{ff}(k_{po}+k_{do}s)}e^{-sT}\\
    h_{21} &= -\dfrac{s+k_{pf}-e^{-sT}k_{ff}(k_{po}+k_{do}s)}{G_f^{-1}s+k_{pf}-e^{-s2T}k_{ff}(k_{po}+k_{do}s)}\\
    h_{22} &= \dfrac{k_{ff}(1-e^{-s2T}k_{fo})s}{G_f^{-1}s+k_{pf}-e^{-s2T}k_{ff}(k_{po}+k_{do}s)}
\end{align}
Notice that all elements of the $H$ matrix share the same denominator. Consider the case where time delay approaches zero ($T\rightarrow0$) and we use unity-gain force feedback ($k_{fo}=1$). Then $h_{22}=0$ and $h_{12}=1$. The terms $h_{11}$ and $h_{21}$ achieve their ideal values of $0$ and $-1$ respectively if and only if $G_f=1$, which would occur if the follower were infinitely fast and perfectly accurate. Of course this is not the case and the transient response will never be exactly unity. Thus, with no time delay and $k_{fo}=1$, the hybrid matrix becomes
\begin{equation}
    H(s) = \begin{bmatrix}
        \dfrac{(k_{po}+k_{do}s)\left(1-G_f^{-1}\right)}{G_f^{-1}s+k_{pf}-k_{ff}(k_{po}+k_{do}s)} & 1 \\ -\dfrac{s(1-G_f^{-1})}{{G_f^{-1}s+k_{pf}-k_{ff}(k_{po}+k_{do}s)}}-1 & 0
    \end{bmatrix}
\end{equation}
However, at steady-state if there is no remaining error, $G_f$ tends to $1$, so $h_{11}=0$ and $h_{21}=-1$. Similarly, if the operator velocity is very small, for example when in contact with a patient during an ultrasound, $v_o\rightarrow 0$, so transparency is achieved at steady state.

The follower's response, $G_f$, cannot be altered, but $k_{po}$, $k_{do}$, $k_{pf}$, and $k_{ff}$ can be tuned to bring $h_{11}$ closer to $0$. Indeed, taking either follower gain, $k_{pf}$ or $k_{ff}$, to infinity drives $h_{11}$ and $h_{21}$ to their ideal values. With very large operator gains, $k_{po}$ and $k_{do}$, $h_{21}\rightarrow0$ and $h_{11}\rightarrow\frac{G_f-1}{k_{ff}G_f}$. However, such large gains are prevented by practical considerations such as haptic device joint torque limits and stability. Adding time delay further degrades the transparency.

\subsection{Control Architectures}
In this subsection, various possible teleoperation architectures based on this model are described. Their stability is analyzed in Section \ref{ssec:stab}.

\subsubsection{2-Channel Teleoperation}
Three 2-channel teleoperation approaches were implemented, differing in which variable is reflected from follower to operator and whether the follower receives local position feedback. In each case, the operator motion is sent to the follower.

\noindent\textit{Position-Position ($C_2=C_3=0$):}\\
In this architecture, the follower position is fed back to the operator, and a force proportional to the position error and its time derivative is applied to the haptic device. This is one of the first bilateral teleoperation schemes to have been developed and has the advantage of not requiring a force sensor at the follower. In this case, however, $k_{fo}=k_{pf}=k_{ff}=0$, which leads to problems with the transparency. In Fig. \ref{fig:teleop}, this is equivalent to setting $C_2=C_3=C_v^f=C_f^f=C_f^h=0$. The hybrid matrix becomes
\begin{equation}
    H(s) = \bmat{e^{-s2T}\left(\dfrac{k_{po}}{s}+k_{do}\right)(G_f-1) & 0 \\ -G_f & 0}
\end{equation}
The obvious problem is that the haptic device force has no relation to the follower force. In practice, the bigger problem is that the follower lags the operator and the returned force is further delayed, so the operator always feels forces resisting their motion which makes operation difficult and frustrating. This is evident in $h_{11}$, which becomes further from $0$ as the follower response $G_f$ departs from unity and as the round-trip time of the communication ($2T$) increases. For this reason, the position-position architecture was implemented and tested once but discarded for very poor performance.

\noindent\textit{Force-Position ($C_3=C_4=0$):}\label{sssec:forcePos}\\
This teleoperation scheme is the na\"ive method of providing force feedback. Quantitatively, the force-position teleoperation sets $k_{po}=k_{do}=k_{pf}=k_{ff}=0$. In Fig. \ref{fig:teleop}, this is equivalent to setting $C_3=C_4=C_v^f=C_f^f=C_v^h=0$. The hybrid matrix then becomes
\begin{equation}
    H(s) = \bmat{0 & e^{-sT}k_{fo} \\ -G_f & 0}\label{eqn:forcePos}
\end{equation}

The time-delayed force feedback is seen in $h_{12}$, and motion tracking is affected by the follower's response. If the tracking is not fast and accurate, it can lead to oscillation due to the delay. Often in practice $k_{fo}<1$ is used to improve stability. Intuitively, this reduces the jerk on the operator when the novice side changes suddenly, e.g., when first touching the patient. Similarly, lowpass filtering the fed-back force injects damping into the system, greatly reducing the oscillatory behaviour. Both, however, decrease transparency. 

Additionally, in human teleoperation the follower can easily misjudge the alignment of the virtual and real tools due to imperfect depth perception as well as partial occlusion of the real tool by the virtual one. With this controller there is no possibility to correct the follower's pose to account for such steady-state errors.

\noindent\textit{Force-Position with Local Position Feedback ($C_3=C_4=0$):}\\
To correct for these steady state errors, the pose of the follower can be measured, and the tracking error fed back locally. In particular, the control law for the follower is then 
\begin{equation}
\dot{x}_v = \dot{x}_o+k_{pf}(x_o-x_f)
\end{equation}
In Fig. \ref{fig:teleop}, this is equivalent to setting $C_3=C_4=C_f^f=C_v^h=0$. To render this on the HoloLens 2, the desired velocity is numerically integrated using the time step between successive frames, and the resulting pose is applied to the virtual probe. To avoid losing track of the virtual probe in case of poor tracking, $x_v$ is limited to stay within a distance $d_{max}$ of $x_o$. By adding the position error to the virtual probe signal, the tracking error is effectively amplified with an integrator. If the follower is not well aligned in one axis, the virtual probe starts moving further away in that axis, causing the follower to realize their error and better align the probe. A deadband of an acceptably small distance error, $d_{err}$, is applied to the integrator since the error is never exactly zero. 

This control scheme creates the following hybrid matrix, with $k_{po}=k_{do}=k_{ff}=0$.
\begin{equation}
    H(s) = \bmat{0 & e^{-sT}k_{fo} \\ -\dfrac{s+k_{pf}}{G_f^{-1}s+k_{pf}} & 0}
\end{equation}
As expected, this is very similar to the position-force architecture, but $k_{pf}$ provides an opportunity to modify the $h_{21}$ element, giving kinematic correspondence.

\subsubsection{3-Channel Teleoperation ($C_4=0$)}
As in the 2-channel force-position teleoperation, only the follower's force is reflected to the operator. However, both pose and force are sent from operator to follower, enabling the follower control law to include a local force feedback term as well. This is useful during contact phases of the teleoperation, in which the follower is to apply a certain desired force. With a relatively stiff environment, a minute change in position constitutes a large change in force, so employing only position and velocity in the control law leads to relatively poor force tracking. Therefore, the follower control law is given by
\begin{equation}
    \dot{x}_v = \dot{x}_o + k_{pf}(x_o-x_f) + k_{ff}(f_h-f_f)
\end{equation}
In Fig. \ref{fig:teleop}, this is equivalent to setting $C_4=C_v^h=0$. The hybrid matrix with this controller becomes
\begin{equation}
    H(s) = \begin{bmatrix}
        0 & e^{-sT}k_{fo} \\ -\dfrac{s+k_{pf}}{G_f^{-1}s+k_{pf}} & \dfrac{k_{ff}(1-e^{-s2T})s}{G_f^{-1}s+k_{pf}}
    \end{bmatrix}\label{eqn:3chH}
\end{equation}
From this it is apparent that the follower force now has a bearing on follower velocity, as desired. Numerically, this decreases transparency. However, in practice the follower is relatively insensitive to small changes in virtual probe pose and the human hand has limited motion resolution. These effects are not well modeled and make the added local force feedback term desirable.

\subsubsection{Model-Mediated Teleoperation}
One problem with the above methods is that the operator's sensation depends directly on the follower's actions. While this gives a true representation of the system on the follower side, it can lead to an unsteady experience for the operator who feels every small jolt and inadvertent movement of the follower. This effect can be reduced using low-pass filtering, but only at the cost of response speed. Furthermore, the performance decreases with increasing time delays. 

One method to potentially overcome both problems is to render a local virtual haptic environment for the operator that is a replica of the follower's real environment. To do so, a point cloud or mesh of the follower environment is captured and sent to the operator. This mesh is used as a virtual fixture or keep-out volume by the haptic device, which applies an outward force when moved into the volume by the operator. This is known as model-mediated teleoperation \cite{mitra2008}. Additionally, the impedance of the follower environment can be estimated and applied to the virtual mesh to replicate the feel of the real environment.

In methods presented and tested in this paper we assume an accurate mesh of the environment is available. This is feasible through RGB-D cameras, LiDAR, time-of-flight (ToF) depth cameras, stereo reconstruction, or even monocular 3D reconstruction from camera motion. Almost all robotic or human teleoperation systems are equipped with at least one of the aforementioned sensing modalities. For example, the HoloLens 2 has stereo cameras and a ToF depth sensor, and can easily be augmented with a higher-performance external RGB-D camera. The local model of the environment is sent to the operator and can be updated if the environment changes. 

In addition, the impedance of the environment is assumed to be estimated accurately in real time by the force and pose tracking of the follower tool. Since the environment impedance is generally relatively constant, temporary loss in communication is less disruptive in this method than those described in previous subsections because the system can continue to render the last known impedance. However, there may also be sudden changes in impedance as the follower moves, so the impedance estimation must converge quickly.

The haptic device and follower forces are given simply by the patient model with the estimated impedance:
\begin{align}
    f_h &= -\hat{b}_p\dot{x}_o-\hat{k}_{p}x_o\label{eqn:fhmesh}\\
    f_f &= -b_p\dot{x}_f-k_{p}x_f \label{eqn:ffmesh}
\end{align}
Plugging in the follower model gives
\begin{align}
    f_h &= -\left(\hat{b}_p+\dfrac{\hat{k}_p}{s}\right)v_o\\
    f_f &= -\left(b_p+\dfrac{k_p}{s}\right)G_fv_v
\end{align}
For the virtual probe pose, we can again use Equation \ref{eqn:lpfLaw}, and substitute in $v_f=G_fv_v$.
\begin{align}
    v_v &= e^{-sT}v_o+\dfrac{k_{pf}}{s}(e^{-sT}v_o-G_fv_v)+k_{ff}(e^{-sT}f_h-f_f)\nonumber\\
    &= \dfrac{1+\dfrac{k_{pf}}{s}-k_{ff}\left(\hat{b}_p+\dfrac{\hat{k}_p}{s}\right)}{1+\left[\dfrac{k_{pf}}{s}-k_{ff}\left(b_p+\dfrac{k_p}{s}\right)\right]G_f}e^{-sT}v_o
\end{align}
Thus we obtain an expression for the follower velocity:
\begin{equation}
    v_f = \dfrac{1+\dfrac{k_{pf}}{s}-k_{ff}\left(\hat{b}_p+\dfrac{\hat{k}_p}{s}\right)}{\dfrac{1}{G_f}+\dfrac{k_{pf}}{s}-k_{ff}\left(b_p+\dfrac{k_p}{s}\right)}e^{-sT}v_o
\end{equation}
Similarly, from Equations \ref{eqn:fhmesh} and \ref{eqn:ffmesh}, the follower force is
\begin{equation}
    f_f = \left(\dfrac{b_p+\dfrac{k_p}{s}}{\hat{b}_p+\dfrac{\hat{k}_p}{s}}\right)\dfrac{1+\dfrac{k_{pf}}{s}-k_{ff}\left(\hat{b}_p+\dfrac{\hat{k}_p}{s}\right)}{\dfrac{1}{G_f}+\dfrac{k_{pf}}{s}-k_{ff}\left(b_p+\dfrac{k_p}{s}\right)}e^{-sT}f_h
\end{equation}

Here it is clear that if $G_f=1$ and the impedance estimate is perfect, the follower motion and force will correspond precisely to the delayed operator motion and force, so the hybrid matrix will be

\begin{equation}
    H = \bmat{0 & e^{sT} \\ -e^{-sT} & 0}\label{eqn:perfDelay}
\end{equation}

Notice how the time delay in the force is negative. Depending on the communication delay, the operator feels the force feedback from the local model before the follower has applied it. In this way, the model-mediated method gives predictive haptic feedback. 

In general, for imperfect impedance estimation, the damping is very small, and at steady-state the velocity should be zero, so damping is negligible. Assuming $G_f$ is constant, the steady state gain is 
\begin{equation}\label{eq:mmtss}
    f_f = \left(\dfrac{k_p}{\hat{k}_p}\right)\dfrac{k_{pf}-k_{ff}\hat{k}_p}{k_{pf}-k_{ff}k_p}f_h
\end{equation}
If $k_{pf}-k_{ff}k_p>0$ and the tissue impedance is underestimated, i.e. $k_p>\hat{k}_p$, the operator moves too far into the patient for a given force, so the follower must press harder to match the virtual tool and $f_f>f_h$. Conversely, overestimation leads to $f_f<f_h$.

If the impedance model is inaccurate, a small $k_{ff}$ reduces the pose tracking error due to the mismatch, but the force experienced by the operator remains proportionally inaccurate. If the follower response, $G_f$ is poor, large $k_{ff}$ can suppress the error. Thus, every effort should be made to have an accurate and fast impedance estimation and patient model while using a reasonably large $k_{ff}$ to compensate for the follower response. Making $k_{ff}$ too large, however, becomes unintuitive for the follower and amplifies modeling errors. Additionally, taking $k_{pf}=0$ ensures a steady-state gain of $1$ but may lead to poor position tracking. Thus, the best strategy may be to have large $k_{ff}$ and zero $k_{pf}$ in the normal direction to the tissue (the direction of the force), and non-zero $k_{pf}$ in the tangent directions.

Other approaches are also possible wherein virtual tool color, error bars, arrows, or other rendering techniques are used to give visual force or pose error feedback to the follower. However, these tend to overwhelm the follower with too much information and should be avoided if possible \cite{black2023ijcars}. These are also difficult to model and are not considered in this paper.

\subsection{Stability}\label{ssec:stab}
Since the follower is a person, not a robot, instability does not lead to violent motions that cause harm or damage. However, it does render the teleoperation completely unusable, so some analysis of stability can give insight into how well the system will perform. In this section we use a state space formulation of the model that is derived in the Appendix.

For the zero time delay case, the system stability is given by the eigenvalues of the $A$ matrix in Equation \ref{eqn:stateSpace}. These depend on the choice of the feedback and feedforward gains from Equations \ref{eqn:lpfLaw1} and \ref{eqn:lpfLaw}. These can lead to stable or unstable behavior and must be chosen carefully. 

For force-position control with no local feedback loops, only $k_{fo}$ is non-zero. In this case, $A$ is stable. However, it has one eigenvalue close to zero. Increasing $k_{fo}$ makes the follower slower and more oscillatory and in turn pushes this eigenvalue closer to zero. At large enough $k_{fo}$, the system becomes unstable. Thus, we expect that with poor tracking from the follower, the force-position feedback will be oscillatory.

Conversely, if we add local position and force feedback for the follower, the hybrid matrix is given in Equation \ref{eqn:3chH}. By the Routh-Hurwitz criterion for a third degree polynomial, $h_{21}$ is stable if 
\begin{equation}
    b_f(k_f+k_{pf}b_f)-m_fk_{pf}k_f>0
\end{equation}
This is clearly true if $k_{pf}=0$, meaning no local position feedback. With the feedback, however, using the parameter values that gave realistic results in the follower model simulations, the system is stable for any $k_{pf}\geq 0$. Depending on the follower's response, however, this may change. Notice also that with no time delay the stability is independent of the local force feedback, $k_{ff}$. 

For the general time delayed case, we express the state space model from Equation \ref{eqn:delayStateSpace} in the Appendix with time delays as follows
\begin{equation}
    \pmb{\dot{x}}(t) = A_0\pmb{x}(t) + \sum_{n=1}^2A_{n}\pmb{x}(t-nT) + Bx_o^*(t)
\end{equation}
The $A_0$, $A_1$, and $A_2$ matrices are derived in the Appendix.

According to Mori et al. and Cheres et al. \cite{mori1981, cheres1989}, the system is asymptotically stable independent of time delays if
\begin{equation}\label{eqn:mori}
    \mu(A_0) + \sum_{n=1}^2||A_{n}||_2 < 0
\end{equation}
Where $\mu(A)=\frac{1}{2}\lambda_{max}(A^\top + A)$ and $||A||_2=\sqrt{\lambda_{max}(A^\top A)}$, where $\lambda_{max}$ is the maximum eigenvalue. This is difficult to show in general, but can be used to test specific known gains and parameters. 

We can also consider the controllers described above using the hybrid matrices. Anderson and Spong (1989) showed that a system is passive if the norm of its scattering operator is less than or equal to 1:
\begin{align}
    S(s) &= \bmat{1 & 0\\0 & -1}(H(s)-I)(H(s)+I)^{-1}\\
    ||S(s)||_2 &\leq 1 \implies \text{Stable}\nonumber
\end{align}

For example, the hybrid matrix in Equation \ref{eqn:perfDelay} leads to a scattering operator norm of $1$, so stability is maintained irrespective of delay. However, this is more difficult to show when the follower response or parameter estimation are non-ideal. In reality, the impedance estimation is time-varying, delayed, and not perfectly accurate. Furthermore, passivity is a very conservative method of guaranteeing stability, so a system may be stable even if $||S(s)||_2>1$. 

In the following experiments, we will show empirically that stability is much less of a concern in human teleoperation than in robotic teleoperation, and is in this case secondary to performance and transparency.

\begin{figure}[h]
    \centering
    \includegraphics[width=\linewidth]{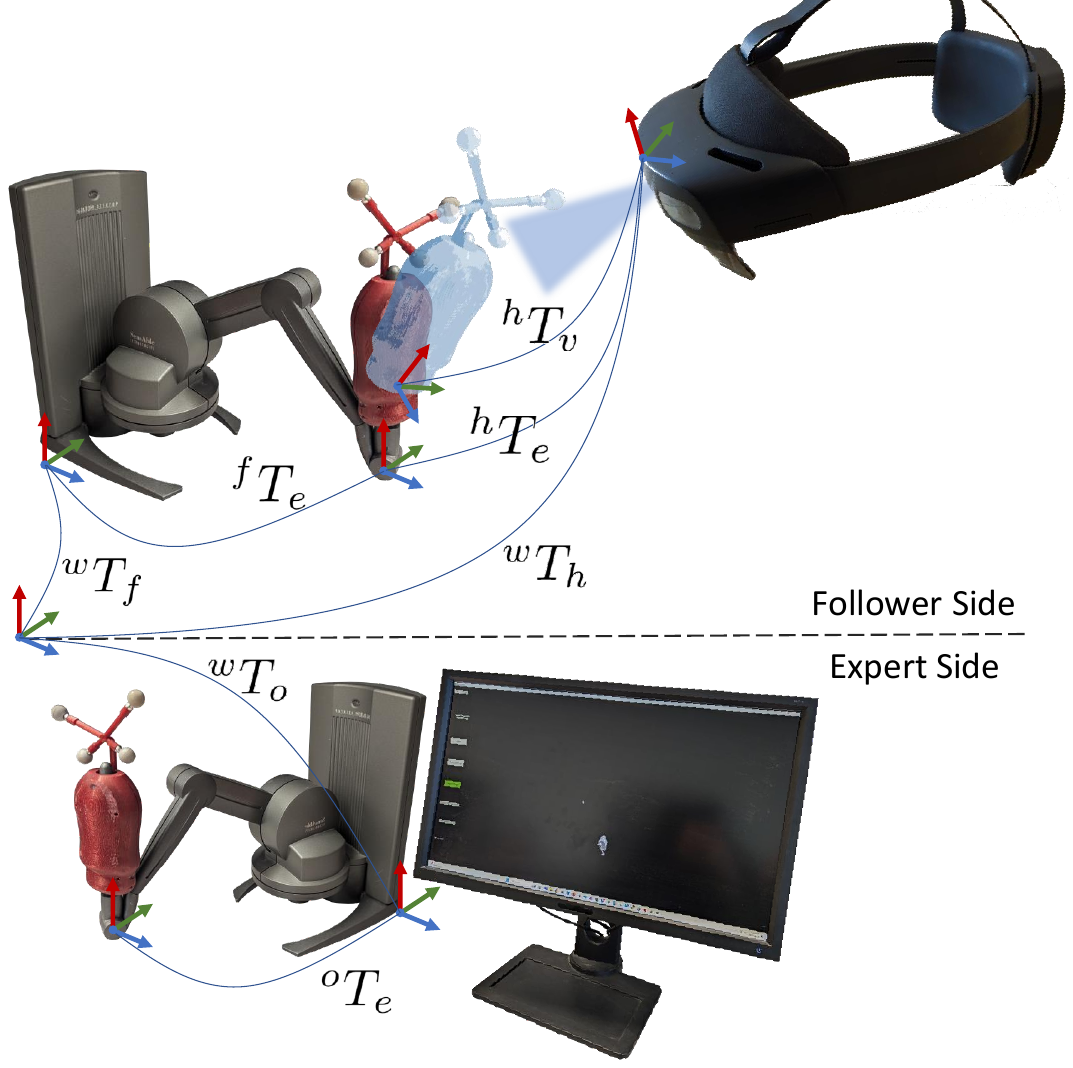}
    \caption{Experimental Setup showing the operator and follower sides which shared one host computer but were separated by a screen. The virtual probe rendered by the HoloLens 2 is shown in semitransparent blue. The red objects on the haptic devices are the ultrasound probe-shaped end effectors with retro-reflective IR markers.}
    \label{fig:expSetup}
\end{figure}

\subsection{Experimental Setup}\label{sec:2f}
To test the human teleoperation architectures in a controlled and close-to-ideal environment, we developed the following system. The follower and operator sides both consisted of a host PC running Windows 11 and a haptic device (Touch X, 3D Systems, Rock Hill, SC). The haptic devices were controlled via C++ programs using OpenHaptics, with a graphical user interface written in Qt. The two clients communicated over a fast WebRTC connection, exchanging position, velocity, and force data, as well as occasional synchronization messages to measure the communication time delay \cite{black2024tmrb}. The time delay was controlled by handling all sending and receiving of WebRTC messages on the operator side on a separate thread, which delayed every message by a set time before sending it or forwarding it to the main graphics and haptics threads. The follower side also had a HoloLens 2 MR HMD which communicated with the follower side PC application over a local WebSocket, receiving messages forwarded from the WebRTC connection.

The virtual tool used in the tests was an ultrasound probe viewed through the HoloLens 2. An identical ultrasound probe shape was 3D printed and attached to the follower's Touch X stylus. Four infrared (IR) reflective spheres were also attached to the dummy probe and were tracked by the HoloLens 2, allowing the HoloLens to compute the haptic device end effector pose in the HoloLens frame \cite{black2024pose}. This was used to perform a registration before every test. Suppose the operator's haptic device has base frame represented by the homogeneous transform ${}^wT_o$ relative to the world while the follower's has base frame ${}^wT_f$ and the HoloLens is at ${}^wT_h$, as shown in Fig. \ref{fig:expSetup}. When the operator moves their haptic device end effector to a pose ${}^wT_o {}^oT_e$, the virtual probe is rendered in the HoloLens at pose ${}^wT_h{}^hT_v$, and the follow moves their handle to pose ${}^wT_f {}^fT_e$. The virtual probe should be rendered such that when the follower aligns perfectly to the virtual probe (i.e. ${}^wT_f {}^fT_e={}^wT_h{}^hT_v$), their handle is at ${}^fT_e={}^oT_e$. In this way, if the follower matches the operator, the poses output by the haptic device control software, OpenHaptics, are equal. Note that the follower pose is ${}^fT_e=({}^wT_f)^{-1}{}^wT_h{}^hT_v$, so to achieve ${}^fT_e={}^oT_e$, the virtual probe pose must be set to ${}^hT_v=({}^wT_h)^{-1}{}^wT_f{}^oT_e$. For this, we must determine the follower haptic device base frame relative to the HoloLens: ${}^hT_f=({}^wT_h)^{-1}{}^wT_f$.

Thus, at the start of each test, the follower haptic device end effector was moved to several different positions and orientations while the HoloLens recorded the measured pose (${}^hT_e$) and received the actual pose from OpenHaptics (${}^fC_e$) over the WebSocket. These are related by ${}^wT_h{}^hT_e={}^wT_f{}^fT_e$, so ${}^hT_f=({}^wT_h)^{-1}{}^wT_f={}^hT_e({}^fT_e)^{-1}$. After recording several hundred samples of $({}^hT_e,~{}^fT_e)$, the least squares-optimal value of ${}^hT_f$ was computed using the Kabsch-Umeyma algorithm, which ensures a valid homogeneous transform using singular value decomposition (SVD) \cite{lawrence2019}. 

During tests, the operator performed arbitrary motions with their haptic device. The motion and force were sent to the follower PC over WebRTC, and forwarded to the HoloLens over the WebSocket. Here they were transformed into HoloLens coordinates and rendered to the follower through the virtual probe pose. The follower grasped the dummy ultrasound probe on their haptic device end effector and aligned it with the virtual probe to follow the desired motion. A virtual environment in the form of a flat surface of specified stiffness and damping was rendered haptically to the follower, leading to forces generated due to the motions. The applied forces and follower motions were sent to the HoloLens for rendering local feedback, as described above, and to the operator for haptic feedback. Depending on the architecture being tested, the operator's haptic device either applied forces based on this feedback and/or simply rendered an identical virtual environment to the follower one to test the model-mediated feedback.

The operator and follower PC clocks were synchronized by sending timestamps along with every data message, and all forces, positions, orientations, and velocities were in the same coordinate frame since both sides used identical haptic devices. All data received on the operator side was recorded to a file for later analysis in MATLAB (Mathworks Inc., Natick, MA).\vspace{-3ex}
 \begin{figure*}[t]
    \centering
    \includegraphics[width=\linewidth]{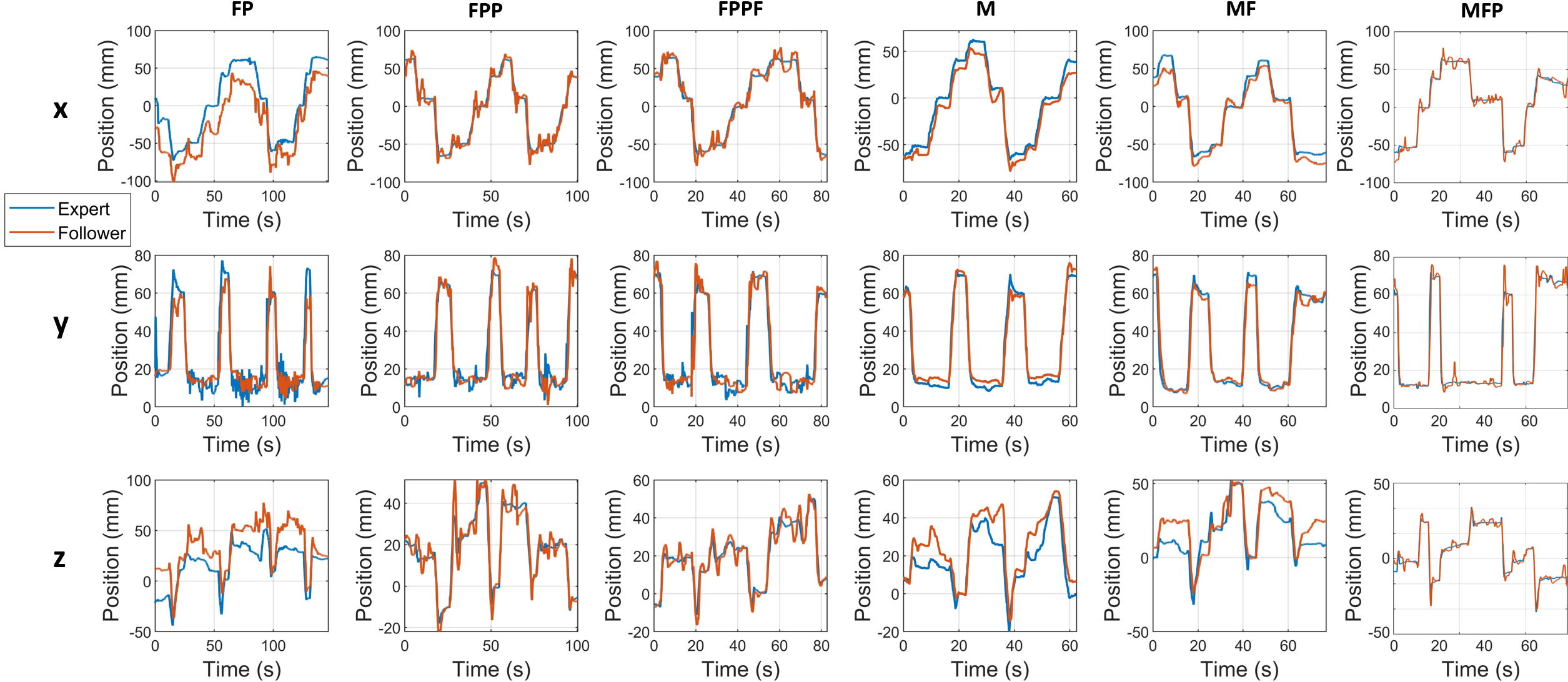}
    \caption{Position tracking with no time delay for the six teleoperation architectures. FP shows significant oscillation during contact while FP, M, and MF have relatively poor position tracking. Adding explicit local position feedback for the follower greatly improves the tracking accuracy and speed.}
    \label{fig:trackingTest}
\end{figure*}
\subsection{Tests}\label{sec:2g}
Using this setup, we performed preliminary tests of the various teleoperation architectures. These tests provide a validation of our mathematical modeling under controlled conditions. The architectures we considered are listed below and described in the previous sections.

\begin{itemize}
    \item Position-Position (PP):\\
    $\dot{x}_v = \dot{x}_o$ and $f_h = k_{po}(x_f-x_o) - k_{do}\dot{x}_o$
    \item Force-Position (FP):\\
    $\dot{x}_v = \dot{x}_o$ and $f_h = k_{fo}f_f - k_{do}\dot{x}_o$
    \item Force-Position with local position feedback (FPP):\\
    $\dot{x}_v = \dot{x}_o + k_{pf}(x_o-x_f)$ and $f_h = k_{fo}f_f - k_{do}\dot{x}_o$
    \item Force-Position with local pose and force feedback (FPPF):\\
    $\dot{x}_v = \dot{x}_o + k_{pf}(x_o-x_f) + k_{ff}(f_h-f_f)$ and \\$f_h = k_{fo}f_f - k_{do}\dot{x}_o$
    \item Model-mediated (M):\\
    $\dot{x}_v = \dot{x}_o$ and\\
    $f_h = \begin{cases}
        k_p(x_0-x_o) - b_p\dot{x}_o & \text{if } x_o \text{ inside mesh}\\
        -b_p\dot{x}_o & \text{otherwise}
    \end{cases} $\\
    where $x_0$ is the position of the virtual mesh surface.
    \item Model-mediated with local position feedback (MP):\\
    Same as M but with\\
    $\dot{x}_v = \dot{x}_o + k_{pf}(x_o-x_f)$
    \item Model-mediated with local position and force feedback (MFP):\\
    Same as M and MP but with\\
    $\dot{x}_v = \dot{x}_o + k_{pf}(x_o-x_f) + k_{ff}(f_h-f_f)$
\end{itemize}

Note, in MFP the force feedback was given normal to the surface and the position feedback was tangent to the surface. The surface in these tests was a flat, horizontal virtual surface in the haptic device with set stiffness and damping. 

For each teleoperation architecture, the operator performed arbitrary smooth motions for 1 to 4 minutes with the follower tracking. The position and force root-mean-squared (RMS) tracking errors were analyzed, as well as the steady state error for step-like motions. Next, the two most promising methods of each type (force feedback and model-based feedback) were selected and tested again five times, each time with increasing communication time delay (0, 50, 250, 500, and 1000 ms). Then the same methods were tested with three different environment stiffness values (0.2, 0.4, and 0.8 N/mm) and no time delay, with and without low-pass filtering.

In the tests, the operator was one of the authors, and the follower was a volunteer. Five followers were tested, including three males and two females with age ranging from 22 to 57, with various backgrounds. None were experienced in ultrasound or MR. Ethics approval for the tests was obtained from the University of British Columbia Behavioural Research Ethics Board (BREB), approval number H22-01195. Informed consent was given by all participants.

\section{Results}\label{sec:res}
\begin{table}[h]
    \centering
    \caption{Teleoperation performance for different control schemes at zero time delay. The RMS position tracking error, $e_p$, steady-state position error, $e_{pss}$, RMS force tracking error, $e_f$, steady state force error, $e_{fss}$, and approximate time delay, $\tau$, between follower and operator are shown.}
    \label{tab:generalPerf}
    \begin{tabular}{|c|c|c|c|c|c|c|}\hline
         & $e_{p}$ & $e_{pss}$ & $e_{f}$ & $e_{fss}$ & $\tau$ \\
         & (mm) & (mm) & (N) & (N) & (ms) \\\hline
       FP & 7.92 & 6.57 & \textbf{0.18} & 0.12 & 440 \\\hline
       FPP & 5.91 & 3.83 & 0.16 & 0.11 & 38 \\\hline
       FPPF & 6.18 & \textbf{1.52} & 0.33 & \textbf{0.07} & 48 \\\hline
       M & 6.75 & 5.12 & 2.66 & 2.45 & 254 \\\hline
       MF & 6.91 & 6.23 & 0.94 & 0.9 & 284 \\\hline
       MFP & \textbf{3.86} & 3.38 & 0.58 & 0.47 & \textbf{36} \\\hline
    \end{tabular}
\end{table}
\subsection{Tracking Tests with no Delay}
\begin{figure}[h]
    \centering
    \includegraphics[width=\linewidth]{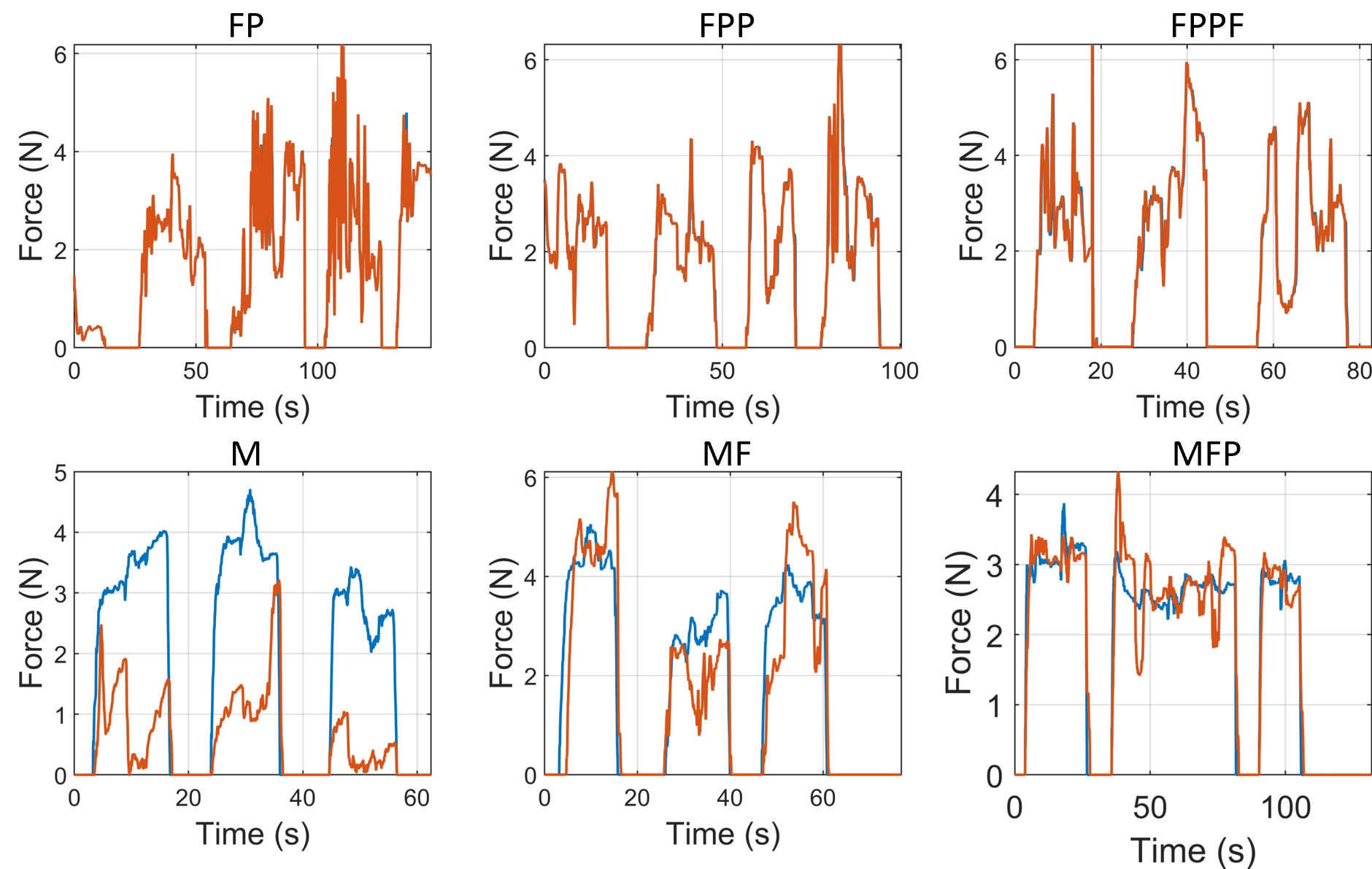}
    \caption{Force tracking with no time delay using various teleoperation schemes.}
    \label{fig:forceTracking}
\end{figure}

The average results of the first test are shown in Table \ref{tab:generalPerf}, and one set of tests is plotted in Figs. \ref{fig:trackingTest} and \ref{fig:forceTracking}. For these nominally zero time delay tests, the communication round trip time was measured to be $3.69\pm 2.67$ ms (average $\pm$ standard deviation) using the method described in \cite{black2024tmrb}. For later non-zero delay tests, the nominal time delay is also reported but may vary similarly.
\begin{figure*}[h]
    \centering
    \includegraphics[width=0.9\linewidth]{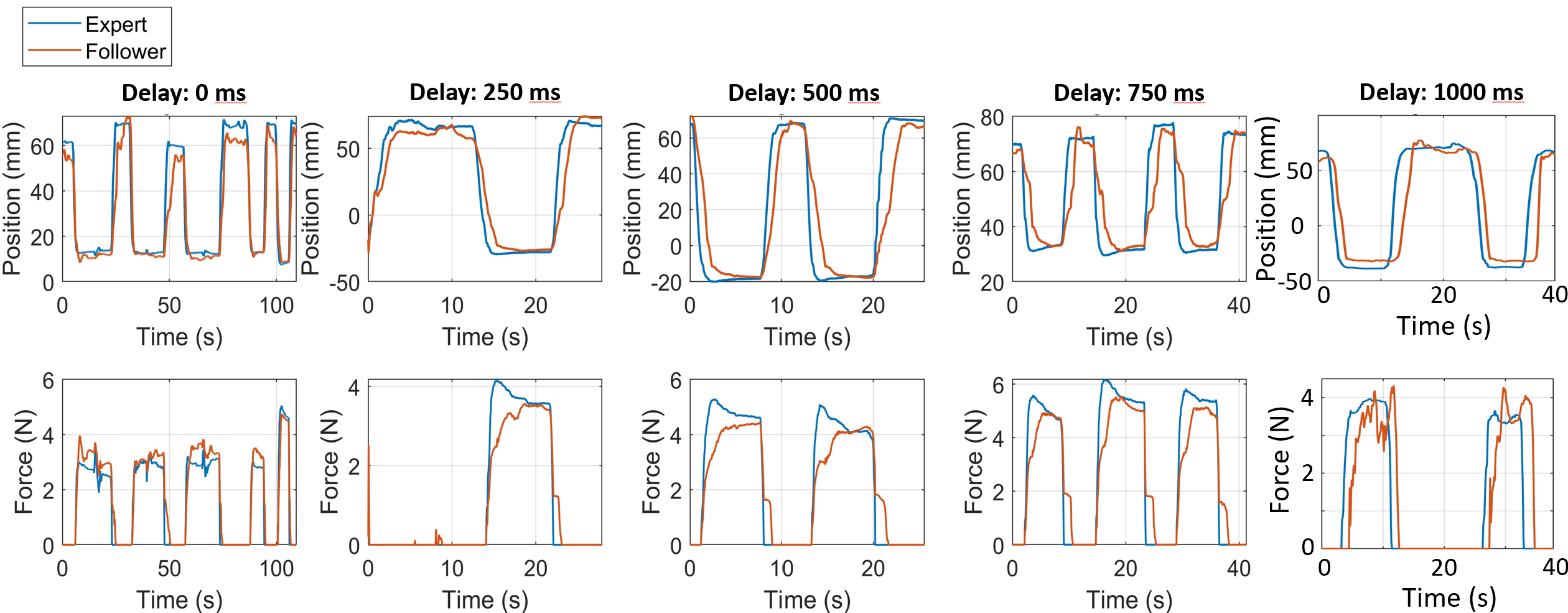}
    \caption{Position and force tracking (both normal to the virtual surface) in the presence of time delays with the MFP controller (mesh local model with local position and force feedback). Time delay has little effect on the force tracking and only delays the position tracking.}
    \label{fig:meshTracking}
\end{figure*}

Even with no delay, it was immediately apparent that position-position teleoperation was impractical. The lag of the follower in tracking the input signal made it difficult for the operator to move at all if the stiffness was high, while low stiffness gave the operator very poor feeling of where the surface was. Additionally, the follower, being human, does not track very smoothly, so the operator continually received unwanted and jarring feedback. Low-pass filtering helped but increased the delay, which again made it harder for the operator to move. Thus, while the position tracking was by definition good, the teleoperation experience was intolerable and the method was removed from the tests.

\begin{figure*}[h]
    \centering
    \includegraphics[width=0.9\linewidth]{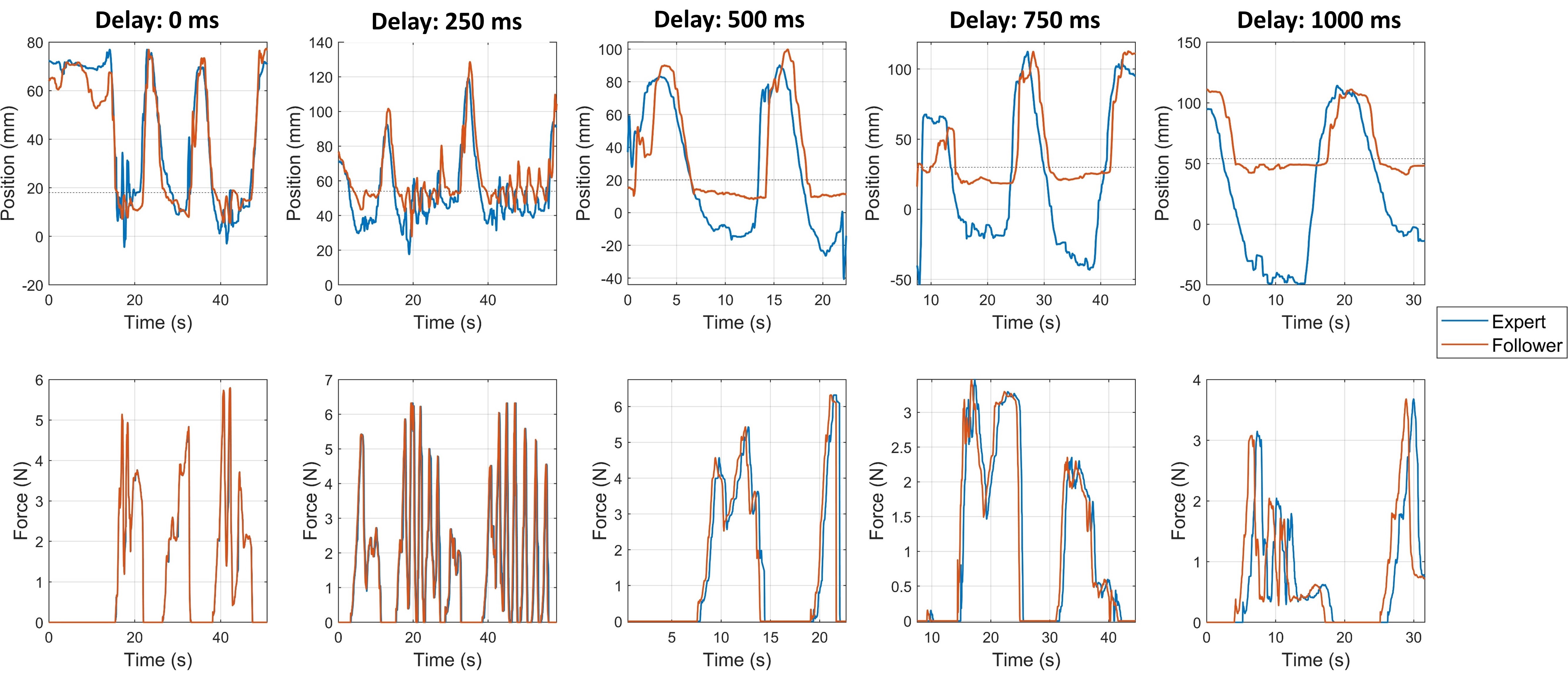}
    \caption{Position and force tracking (both normal to the virtual surface) in the presence of time delays with the FPPF controller (force-position teleoperation with local position and force feedback for the follower). With delays over approximately 500 ms, tracking becomes infeasible.}
    \label{fig:posDelay}
\end{figure*}
Direct force feedback performed much better, though the lack of explicit position coordination on the operator side was apparent, with relatively large tracking errors. Additionally, contact was oscillatory: due to the tracking lag of the follower, the operator initially moved well below the surface before receiving force feedback which was stronger than anticipated due to the follower's attempt to reach the virtual tool inside the surface. This jerked the operator's hand upward which the follower tried to track, thus decreasing the force, so the operator's hand moved down again. This leads to periodic and uncomfortable motion when in intermittent contact. This is similar to the chattering effect in robots, but at a lower rate and larger amplitude. This back-and-forth reflection of waves is very apparent in Fig. \ref{fig:forceTracking}. Both behaviors match what was expected from Equation \ref{eqn:forcePos}. 

Adding local position feedback for the follower helped decrease position error and substantially reduced tracking lag. By effectively magnifying any tracking error, the local feedback has a predictive effect, moving the probe further than the operator has yet moved and encouraging the follower to react immediately. This reduced the oscillation during contact. The feedback also led to much more accurate position tracking, as shown in Fig. \ref{fig:trackingTest}. With local position and force feedback, position tracking suffered slightly, since the virtual probe no longer exclusively represents the desired pose, but aims to control the force as well. In Equation \ref{eqn:3chH}, this is reflected by the non-zero $h_{22}$ element, while $k_{pf}$ enables the more responsive position tracking. Interestingly, RMS force tracking also became slightly worse since the increased feedback to the follower sometimes caused initial overshoot. However, in steady-state the force quickly converged to be more accurate than without local feedback. The addition of force additionally had a damping effect during contact, discouraging sudden changes in force and thus leading to much less oscillation.

The teleoperation architectures based on a local model or mesh showed different behavior. Contact oscillation was completely eliminated due to the stable and unchanging mesh. This led to a much more comfortable and intuitive experience. With local position feedback, the position tracking was the best of any of the methods. On the other hand, force tracking was completely lacking unless explicitly enforced with local feedback, as shown in Fig. \ref{fig:forceTracking}. Even so, the force experienced by the operator was not as accurate as when the measured force was fed back directly, and is affected by modeling errors in the surface shape and impedance. Again, adding local position feedback greatly reduced the tracking lag.

\subsection{Tracking Tests with Delay}
The effect of communication delays on the force and position tracking is shown in Figs. \ref{fig:meshTracking} and \ref{fig:posDelay} for the FPPF and MFP controllers. Due to the delay, discussing RMS error is not meaningful. However, several important results are apparent in the plots. With the 3-channel architecture, the character of the response underwent two distinct phases. Initially, up to between 250-500 ms delay, the response became increasingly oscillatory as the wave reflections were exacerbated by the larger delays. At 500 ms, however, the operator's approach changed and they started moving more slowly and carefully, anticipating the unexpected application of force feedback. As a result, the oscillations ceased. However, as the delay continued increasing, the operator was deeper and deeper within the surface before the follower's force was applied, measured, and fed back. This led to excessive offsets between the virtual probe and the follower, making tracking impossible. 

Conversely, with model-mediated teleoperation, the operator always received immediate haptic feedback with no delay. Consequently, their motion was unaffected and the follower was able to track without problems, simply delayed by the given amount. Thus, while the delay made teleoperation impractical with direct force feedback, it had little effect when operating with a local model. Due to the local position and force feedback on the follower side, the tracking was good for both. This reflects what was expected in Equation \ref{eqn:perfDelay} with perfect modeling, but will degrade with an imperfect local model.

\subsection{Environment Stiffness and Low-pass Filtering}
\begin{figure*}[h!]
    \centering
    \includegraphics[width=\linewidth]{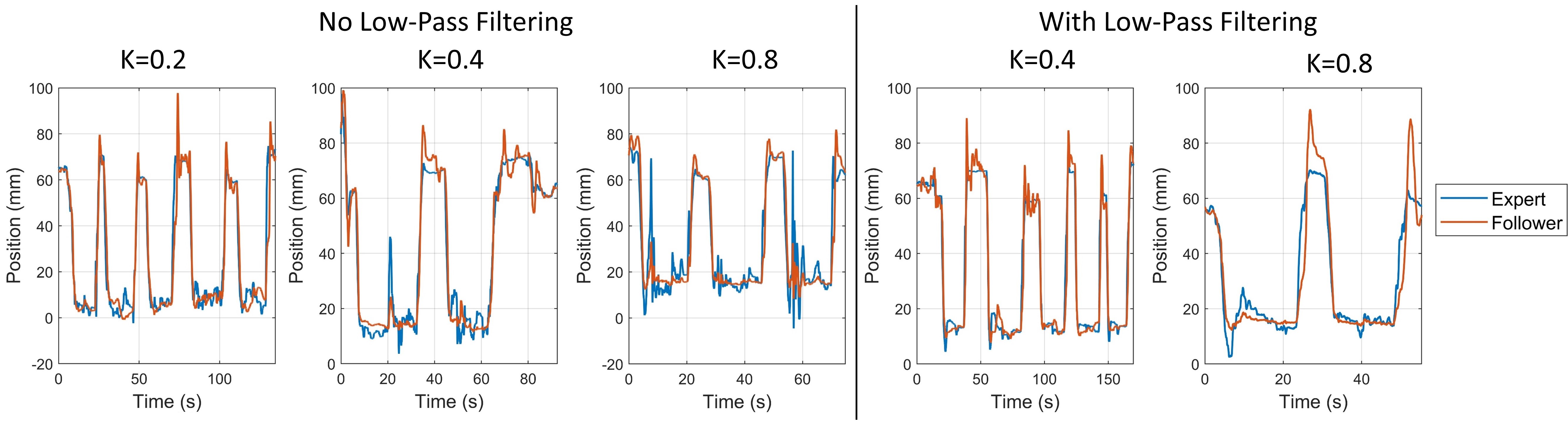}
    \caption{Position tracking at different environment stiffnesses with and without low-pass filtering, with the FPPF teleoperator. In contact (position close to 0 mm), the oscillations increase with stiffness but decrease with low-pass filtering.}
    \label{fig:stiffLP}
\end{figure*}

Finally, the effect of environment stiffness on the teleoperation performance was tested. At zero delay, the stiffness was varied from $0.4$ N/mm (the value used for all other tests) down to $0.2$ N/mm to simulate a very soft environment and up to $0.8$ N/mm for a stiff environment. The damping was kept constant and was not noticeable. These tests were performed with the FPPF architecture since the stiffness has little effect on the local mesh feedback as long as it is approximately equal for the operator and follower. 

\begin{table}[h]
    \centering
    \caption{Teleoperation performance (RMS tracking error in mm) for stiff and soft environments, with and without low-pass filtering (LP).}
    \label{tab:stiff}
    \begin{tabular}{|c|c|c|c|}\hline
      Stiffness (N/mm) & 0.2 & 0.4 & 0.8 \\\hline
       FPPF & 6.48 & 6.18 & 7.53 \\\hline
       FPPF + LP & 5.04 & 4.78 & 6.31 \\\hline
    \end{tabular}
\end{table}
It was already noted in previous tests that the direct haptic feedback approaches had oscillatory behaviour in contact due to the delay of the follower. Therefore, low-pass filtering of the fed-back force was also tested, using a single-pole infinite impulse response filter with a time constant of approximately 0.1 seconds. This injects damping into the system and is expected to decrease the oscillation.

The results of these tests are shown in Table \ref{tab:stiff} and Fig. \ref{fig:stiffLP}. The oscillatory behaviour increases with stiffness, as expected, but strongly decreases in the presence of low-pass filtering, without greatly increasing the phase lag. This leads to better RMS error, but more importantly substantially improved the feel for the operator. The feedback was much more stable and less disruptive than without the filtering. The lower environment stiffness slightly increased tracking error compared to normal. This is because on the softer virtual surface, the operator and follower could move more for small changes in force, making it harder to maintain a precise position.

\subsection{Robustness to Parameter Variation}\label{ssec:rob}
Throughout this paper, we have made a number of simplifying assumptions about the system. In particular, the follower and operator models are taken to be constant. However, they may be affected by external, time-varying influences such as distractions or fatigue and, as described in Section \ref{sssec:forcePos}, the follower may have imperfect depth perception. Additionally, the follower is often hesitant or unable to apply large forces on a patient, so may fail to follow a virtual tool placed too far within the patient. In this case, $f_f\neq -k_px_f-b_p\dot{x}_f$, as assumed in Fig. \ref{fig:models}, so additional dynamics are introduced. The linear time-invariant assumption in the model is thus not always accurate, and there is both parametric and dynamic uncertainty in the follower and operator models. Thus, the mathematical modeling can be improved through stochastic approaches as both users' actions are not deterministic and their imperfect sensory perception and motor skills lead to random fluctuations in the motions. These imperfections could be treated as disturbances in the existing model in future work. However, as shown in Fig. \ref{fig:modelFit} and the related fit error, the models with constant parameters were able to capture complex recorded motions for an extended period of time without any drift due to parameters changing. Indeed, with controller gains $K_{fo}$, $K_{of}$, $K_{oo}$, and $K_{ff}$ chosen to give realistic performance at the parameter values identified in Table \ref{tab:params}, the acceptable ranges of the model parameters are shown in Table \ref{tab:robstab}. To compute these values, the structured singular value, $\mu$, of the closed-loop system was computed in MATLAB with uncertainty on one parameter at a time. On average, the parameters can individually vary approximately $75\%$ without causing instability. Thus, the LTI assumption gives a reasonable representation of the system.
\begin{table}[h]
    \centering
    \caption{Upper and lower bounds of parameter values to maintain stability. The average acceptable percent variation from the mean parameter value is $74.7\%$ for the system to remain stable.}
    \begin{tabular}{|c|c|c|c|}
        \hline
        Parameter & Minimum & Maximum & Percent Variation\\\hline
        $k_p$ & 4.14 & 15.9 & $58.6\%$ \\
        $b_p$ & 0.093 & 1.91 & $90.7\%$ \\
        $b_f$ & 0.207 & 0.571 & $46.8\%$ \\
        $m_f$ & 0.020 & 0.056 & $47.4\%$ \\
        $m_o$ & 0.000 & 0.2 & $100\%$+ \\
        $b_o$ & 0.000 & 36.5 & $100\%$+ \\
        $k_a$ & 0.002 & 200 & $100\%$\\
        $k_{o1}$ & 0.000 & 1.02 & $100\%$ +\\
        $k_{o2}$ & 63.7 & 96.3 & $20.4\%$\\\hline
    \end{tabular}
    \label{tab:robstab}
\end{table}

Additionally, though it is assumed that the follower can handle intermittent patient contact implicitly without affecting their tracking, switching contact affects the force fed back to the operator and may thus destabilize the closed-loop system. This is shown in Fig. \ref{fig:posDelay}, where the tracking is unstable at 250 ms delay during intermittent contact, leading to very ineffective teleoperation. This shows that instability in human teleoperation does not constitute a health or equipment hazard but makes guidance impossible. However, at higher delays this instability disappears due to an apparent change in strategy. The instability is also eliminated when using model-mediated teleoperation. 

\section{Discussion}\label{sec:discussion}
This paper has introduced the concept of bilateral human teleoperation as a control problem analogous to robotic bilateral teleoperation but with some unique characteristics. The modeling and simulation of the system were described, and various control architectures were explored in terms of their implementation, theoretical transparency and stability, simulation, and practical performance on a test system. The results show that high performance, stable, transparent bilateral human teleoperation is possible but that time delays pose a challenge. With less than approximately 200 ms communication latency, a three channel teleoperation system performed best. In this scheme, the follower's set point velocity is a function of the operator's velocity and the follower's position and force tracking error, and the follower's force is fed back to the operator with significant low-pass filtering. With time delays, however, direct feedback becomes impractical and alternate approaches are required. One such approach of model-mediated teleoperation, in which the operator interacts with a local virtual model of the follower's environment, had excellent performance irrespective of time delay, though it is dependent on an accurate model. 

The paper represents an initial exploration of a novel topic and as a result has some limitations. Firstly, the tests were performed with few subjects and in idealized conditions. They will have to be augmented by practical tests with numerous subjects. For example, to achieve the performance of the ideal model-mediated teleoperation in this paper, future work will have to integrate a depth camera and fast, accurate environment impedance estimation into the system. Both exist, so the assumptions in this paper are not inaccurate. However, very fine details such as the patient's ribs during an ultrasound procedure may have a substantial effect on the measured force and will likely not be captured well in the mesh. 

Similarly, the controllers presented here use pose and force, which have to be measured in real scenarios where the follower is not moving a haptic device. For human teleoperation, the force sensing must be low-profile and have sufficient load capability \cite{black2024ft,black2023whc}, while pose tracking presents another set of challenges. Optical trackers are accurate but susceptible to occlusion, while electromagnetic trackers are affected by external disturbances. Extensive research has been performed in this direction \cite{black2024pose}, but the solutions are imperfect and future work will have to test practical bilateral teleoperation with all of these challenges.

Other future work includes investigation of further control architectures, for example predictive control using the mathematical models to extrapolate trajectories, or robust control by defining uncertainties on the parameters in the derived state space model. This paper also did not consider architectures with non-zero $k_{po}$, i.e. four-channel approaches where the follower position is fed back to the operator. It has been shown that using all four channels of velocity and force is important for optimal transparency \cite{hashtrudi2001}. However, with the follower's unpredictable motions this may not be true for human teleoperation. Traditional passivity-based methods such as wave variables \cite{niemeyer1991} for achieving time delay-robust stability were not explored in this paper because they trade off performance to maintain stability for the sake of safety. As instability in human teleoperation degrades performance rather than safety, this trade-off is not helpful. Time domain passivity control \cite{ryu2005} is less conservative than wave variables and may lead to better performance but is left for future work.

Finally, orientation tracking was not explicitly evaluated in this paper but can be taken to be analogous to position tracking - i.e. consisting of similar lag and accuracy. This was the case in previous tests of human teleoperation \cite{black2023ijcars}. Moreover, in \cite{black2023ijcars}, different rendering schemes were tested. These tests could be extended to find the best rendering method for improved depth perception. During the step-like motions in this paper, the distribution of the follower position was approximately Gaussian about the desired location. However, in the depth direction from the follower's point of view (approximately $z$ in Fig. \ref{fig:trackingTest}), the position variance was significantly greater than in the other directions, pointing to possible depth perception limitations.

In model-mediated teleoperation, the patient impedance model affects the stability. We assumed in the tests that the patient model is fixed and known, with no delay and no error in the parameter estimation. Of course, in reality the patient breathes and moves, and the impedance varies when examining different anatomies. For example, when scanning the liver and kidneys, the probe moves from ribs to soft tissue and back. Practical impedance estimation schemes based on pose and force measurement require several measurement samples to converge to an estimate. Thus, the impedance model is delayed, varying, and not perfectly accurate. However, as shown in Equation \ref{eq:mmtss}, the follower in the model-mediated teleoperation can converge to the correct normal force and tangent position despite modeling errors by having local force feedback as well as local position feedback in directions orthogonal to the force. In this case only the operator feels a slightly incorrect force that is only transient while the impedance estimation converges. In practice, the ultrasound probe is moved slowly and contact is maintained, so the impedance is approximately constant. Thus, the test results are realistic and show the effectiveness of model-mediated teleoperation in this system.

\section{Conclusion}
This paper is a first exploration of bilateral human teleoperation from a controls perspective. The results show that three channel teleoperation is effective at small time delays while model-based teleoperation with local force and pose feedback achieves good performance even with large latency and imperfect impedance estimation. With the developed system model, simulation, and practical setup, future work can build off the presented methods and tests to achieve high-performance, transparent, stable human teleoperation despite time delays. This will in turn enable effective remote guidance and execution of important tasks such as ultrasound exams with the relatively simple, low-cost mixed reality human teleoperation system.

\bibliographystyle{ieeetr}
\bibliography{refs}

\section*{Appendix}
In this section, we derive a state space model of the full teleoperation system. This formulation is used in the paper to assess stability and robustness.

From Section \ref{sec:2b}, the follower is given by
\begin{align}\label{eqn:folss}
    \pmb{\dot{x}}_f=\bmat{\dot{x}_f\\\ddot{x}_f} &= \bmat{0&1\\-\frac{k_{f}}{m_f}&-\frac{b_f}{m_f}}\bmat{x_f\\\dot{x}_f}+\bmat{0&0\\\frac{k_{f}}{m_f}&\frac{b_f}{m_f}}\bmat{x_v\\\dot{x}_v}\nonumber\\
    \pmb{y}_f=\bmat{x_f\\\dot{x}_f\\f_f}&=\bmat{1&0\\0&1\\-k_{p}&-b_p}\bmat{x_f\\\dot{x}_f}
\end{align}
where $k_{p}$, $b_p$ are zero when not in contact with the patient. For this to be LTI, we assume the patient is unmoving and their impedance is constant. This is approximately true when scanning only one region, for example the abdomen. An advantage of this model is that we do not have to consider switching contact, which is handled implicitly by the follower.

\begin{figure*}[t]
    \centering
    \includegraphics[width=\linewidth]{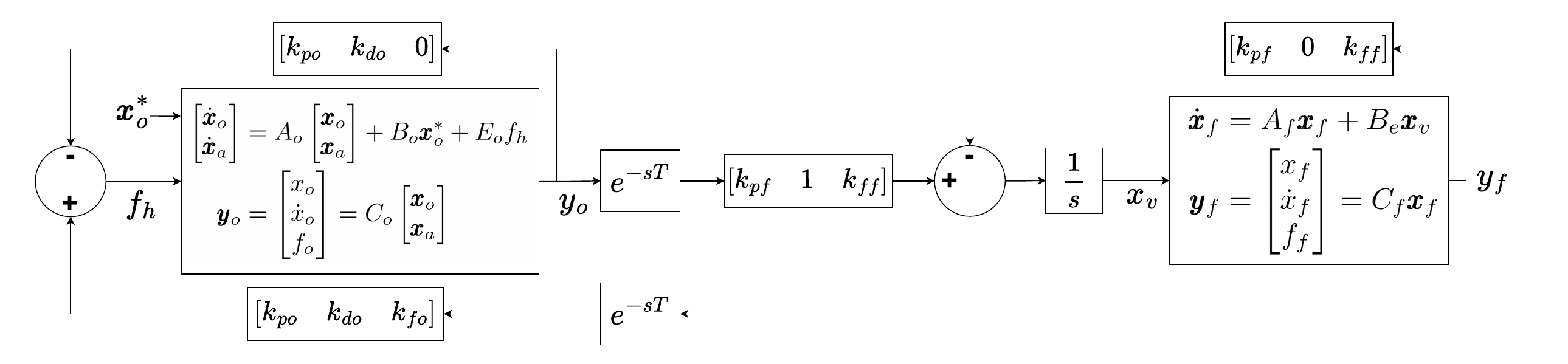}
    \caption{State space representation of teleoperation system. Different behavior is achieved by changing the feedback and feed-forward gains of the positions, velocities, and forces. The operator-side model is given in Equation \ref{eqn:expss} while the follower side is in Equation \ref{eqn:folss}.}
    \label{fig:stateSpaceTeleop}
\end{figure*}

The operator model is given by
\begin{equation}
    m_o\ddot{x}_o+b_o\dot{x}_o=f_h+f_o
\end{equation}
The operator changes their applied force depending on the haptic device force, to track a desired trajectory. In particular, to choose a representative response, we can represent the system in state space and use servo control with pole placement, choosing
\begin{equation}
    f_o=-K_o\pmb{x}_o+k_a\int(x_o^*-x_o)dt
\end{equation}
where $\pmb{x}_o=[x_o,~\dot{x}_o]^\top$, and $x_o^*$ is the operator's desired motion. The state feedback gains are $K_o=\bmat{k_{o1}&k_{o2}}$. These can be tuned to obtain a response that resembles the recorded operator data. We ignore the operator's desired velocity for simplicity as it introduces a zero eigenvalue and the operator is primarily interested in position. Setting $x_a=\int(x_o^*-x_o)dt$, we find the following model
\begin{align}\label{eqn:expss}
    \pmb{\dot{x}}_O=\bmat{\dot{\pmb{x}}_o\\\dot{x}_a} &= \bmat{0&1&0\\-\frac{k_{o1}}{m_o}&\frac{-b_o-k_{o2}}{m_o} & \frac{k_{a}}{m_o}\\-1&0&0}\bmat{\pmb{x}_o\\x_a}\nonumber\\
    &+\bmat{0\\0\\1}x_o^* + \bmat{0\\\frac{1}{m_o}\\0}f_h\nonumber\\
    \pmb{y}_o = \bmat{x_o\\\dot{x}_o\\f_o} &= \bmat{1 & 0 & 0\\0&1&0\\-k_{o1}&-k_{o2}&k_{a}}\bmat{\pmb{x}_o\\x_a}
\end{align}
where the gain on $f_h$ is denoted $E_o$.

\subsubsection{Teleoperation System}
The full teleoperation system is shown in Fig. \ref{fig:stateSpaceTeleop}, given the state space matrices derived above. We assume the follower sees only the virtual tool pose and the operator receives feedback only by forces applied to the haptic device. This does not model the operator's visual feedback from video or other sensor streams such as ultrasound images, nor the verbal communication between operator and follower. Instead, the operator's decisions based on the ultrasound images determine their desired trajectory, $x_o^*$, which is the input to the model. In this case, we control the virtual tool pose, $x_v$, for the follower, and the haptic device force, $f_h$ for the operator. Each can be a function of the position, velocity, and force of the operator and follower as shown in the diagram. Some gains are never used, so they are set to zero. For example, the haptic device is not equipped with a force sensor, so its actual force cannot be fed back. Thus, the general inputs to the operator and follower are
\begin{align}
    f_h &= k_{fo}f_f + k_{po}(x_f-x_o) + k_{do}(\dot{x}_f-\dot{x}_o)\label{eqn:lpfLaw1}\\
    \dot{x}_v &= \dot{x}_o + k_{pf}(x_o-x_f) + k_{ff}(f_h-f_f)\label{eqn:lpfLawapx}
\end{align}
Let $K_{oo}=\bmat{k_{po}&k_{do}&0}$, $K_{of}=\bmat{k_{pf}&1&k_{ff}}$, $K_{ff}=\bmat{k_{pf}&0&k_{ff}}$, and $K_{fo}=\bmat{k_{po}&k_{do}&k_{fo}}$. Then in terms of the state space outputs, these expressions become
\begin{align}\label{eqn:fbeapx}
    f_h &= K_{fo}\pmb{y}_f - K_{oo}\pmb{y}_o\\\label{eqn:fbfapx}
    \dot{x}_v &= K_{of}\pmb{y}_o - K_{ff}\pmb{y}_f
\end{align}
as in Equation \ref{eqn:fbe}. Let $\pmb{x}_v=\bmat{x_v&\dot{x}_v}^\top$ be the state of the virtual tool. The state equations of the operator, follower, and virtual tool are then
\begin{align}
    \dot{\pmb{x}}_O&= A_o\pmb{x}_O+B_ox_o^*+E_o(K_{fo}C_f\pmb{x}_f - K_{oo}C_o\pmb{x}_O)\nonumber\\
    &= (A_o-K_{oo}C_o)\pmb{x}_O+E_oK_{fo}C_f\pmb{x}_f+B_o x_o^*\\
    \dot{\pmb{x}}_f&= A_f\pmb{x}_f+B_{f1}x_v+B_{f2}(K_{of}C_o\pmb{x}_{E}-K_{ff}C_f\pmb{x}_{f})\label{eqn:xfState}\\
    \dot{x}_v &= K_{of}C_o\pmb{x}_O - K_{ff}C_f\pmb{x}_f
\end{align}
Where $B_f=\bmat{B_{f1} & B_{f2}}$. We now define a combined state $\pmb{x}=\bmat{\pmb{x}_O^\top & \pmb{x}_f^\top & x_v}^\top$. This allows us to state the system as one equation:
\begin{align}
\pmb{\dot{x}} &=
    \bmat{A_o-E_oK_{oo}C_o & E_oK_{fo}C_f & 0_{3\times1}\\ B_{f2}K_{of}C_o & A_f-B_{f2}K_{ff}C_f & B_{f1}\\ K_{of}C_o & -K_{ff}C_f & 0}\pmb{x}\nonumber\\
    &+ \bmat{1\\0_{5\times1}}x_o^*\nonumber\\
    \pmb{y}&= \bmat{C_o & 0_{3\times2} & 0_{3\times1}\\0_{3\times3} & C_f & 0_{3\times1}}\pmb{x}\label{eqn:stateSpace}
\end{align}
The dimensions of the matrices are $A\in\mathbb{R}^{6\times 6}$, $B\in\mathbb{R}^{6\times1}$, and $C\in\mathbb{R}^{6\times6}$. 

This analysis has only considered the case of time delay $T=0$. For non-zero delay, we must rewrite Equations \ref{eqn:fbeapx} and~\ref{eqn:fbfapx}:
\begin{align}\label{eqn:fhtapx}
    f_h &= e^{-sT}K_{fo}\pmb{y}_f - K_{oo}\pmb{y}_o\\\label{eqn:xvtapx}
    \dot{x}_v &= e^{-sT}K_{of}\pmb{y}_o - K_{ff}\pmb{y}_f
\end{align}
The system model with time delays is thus
\begin{align*}
    \pmb{\dot{x}}_O &= A_o\pmb{x}_O+B_ox_o^*+E_oe^{-sT}f_h\\
    \pmb{\dot{x}}_f &= A_f\pmb{x}_f+B_{f1}x_v+B_{f2}\dot{x}_v\\
    x_v &= K_{of}C_oe^{-sT}\pmb{x}_O-K_{ff}C_f\pmb{x}_f
\end{align*}
Combining this with equations \ref{eqn:fhtapx} and \ref{eqn:xvtapx}, and with a slight abuse of notation to include time delays in the matrices, we obtain
\begin{align}
\pmb{\dot{x}} &=
    \bmat{A_o-E_oK_{oo}C_oe^{-sT} & E_oK_{fo}C_fe^{-s2T} & 0_{3\times1}\\ B_{f2}K_{of}C_oe^{-sT} & A_f-B_{f2}K_{ff}C_f & B_{f1}\\ K_{of}C_oe^{-sT} & -K_{ff}C_f & 0}\pmb{x}\nonumber\\
    &+ \bmat{1\\0_{5\times1}}x_o^*\nonumber\\
    \pmb{y}&= \bmat{C_o & 0_{3\times2} & 0_{3\times1}\\0_{3\times3} & C_f & 0_{3\times1}}\pmb{x}\label{eqn:delayStateSpace}
\end{align}
Thus, we have the following $A$ matrices:
\begin{align}
    A_0 &= \bmat{A_o & 0_{3\times2} & 0_{3\times1}\\ 0_{2\times3} & A_f-B_{f2}K_{ff}C_f & B_{f1}\\ 0_{1\times3} & -K_{ff}C_f & 0}\nonumber\\
    A_1 &= \bmat{-E_oK_{oo}C_o & 0_{3\times2} & 0_{3\times1}\\ B_{f2}K_{of}C_o & 0_{2\times2} & 0_{2\times1}\\ K_{of}C_o & 0_{1\times2} & 0}\nonumber\\
    A_2 &= \bmat{0_{3\times3} & E_oK_{fo}C_f & 0_{3\times1}\\ 0_{2\times3} & 0_{2\times2} & 0_{2\times1}\\0_{1\times3} & 0_{1\times2} & 0}
\end{align}
These can be used with the Mori and Cheres identity in Equation \ref{eqn:mori} to determine the stability independent of time delays. 
\end{document}